\documentclass{article}

\PassOptionsToPackage{numbers,sort,compress}{natbib}

 \usepackage[preprint]{neurips_2026}

\usepackage[utf8]{inputenc} 
\usepackage[T1]{fontenc}    
\usepackage{hyperref}       
\usepackage{url}            
\usepackage{booktabs}       
\usepackage{amsfonts}       
\usepackage{nicefrac}       
\usepackage{microtype}      
\usepackage{xcolor}         

\usepackage{graphicx}
\usepackage{tabularx}
\usepackage{array}
\usepackage[export]{adjustbox} 

\usepackage[accsupp]{axessibility}  

\usepackage{orcidlink}
\usepackage{multirow} 
\usepackage{makecell}

\title{RoboWM-Bench: A Benchmark for Evaluating \\ World Models in Robotic Manipulation}

\author{%
  Feng Jiang$^{1}$\thanks{Equal contribution} \quad 
  Yang Chen$^{1}$\footnotemark[1] \quad 
  Kyle Xu$^{1}$\footnotemark[1] \quad 
  Yuchen Liu$^{1}$ \quad 
  Haifeng Wang$^{1}$ \quad 
  Zhenhao Shen$^{1}$ \\
  \textbf{Jasper Lu$^{1}$ \quad 
  Shengze Huang$^{2}$ \quad 
  Yuanfei Wang$^{1}$ \quad 
  Chen Xie$^{3}$ \quad 
  Ruihai Wu$^{1,}$\thanks{Corresponding author}} \\
  $^{1}$Peking University \quad $^{2}$Tsinghua University \quad $^{3}$Lightwheel \\
  \texttt{wuruihai@pku.edu.cn}
}

\begin{document}

\maketitle


\vspace{-2mm}
\begin{abstract}
\vspace{-1mm}

Recent advances in large-scale video world models have enabled increasingly realistic future prediction, raising the prospect of using generated videos as scalable supervision for robot learning. However, for embodied manipulation, perceptual realism alone is not sufficient: generated interactions must also be physically consistent and executable by robotic agents. Existing benchmarks provide valuable assessments of visual quality and physical plausibility, but they do not systematically evaluate whether predicted behaviors can be translated into executable actions that complete manipulation tasks.
We introduce \textbf{RoboWM-Bench}, a manipulation-centric benchmark for embodiment-grounded evaluation of video world models. RoboWM-Bench converts generated human-hand and robotic manipulation videos into embodied action sequences and validates them through execution in physically grounded simulation environments. Built on real-to-sim scene reconstruction and diverse manipulation tasks, RoboWM-Bench enables standardized, reproducible, and scalable evaluation of physical executability.
Using RoboWM-Bench, we evaluate state-of-the-art video world models and observe that visual plausibility and embodied executability are not always aligned. Our analysis highlights several recurring factors that affect execution performance, including spatial reasoning, contact prediction, and non-physical geometric distortions, particularly in complex and long-horizon interactions. These findings provide a more fine-grained view of current model capabilities and underscore the value of embodiment-aware evaluation for guiding physically grounded world modeling in robotic manipulation. The project page is available at \url{https://robowm-bench.github.io/RoboWM-Bench/}.

\end{abstract}

\section{Introduction}
\label{sec:intro}
\vspace{-1mm}

\begin{figure*}[t]
\centering
\includegraphics[width=\linewidth]{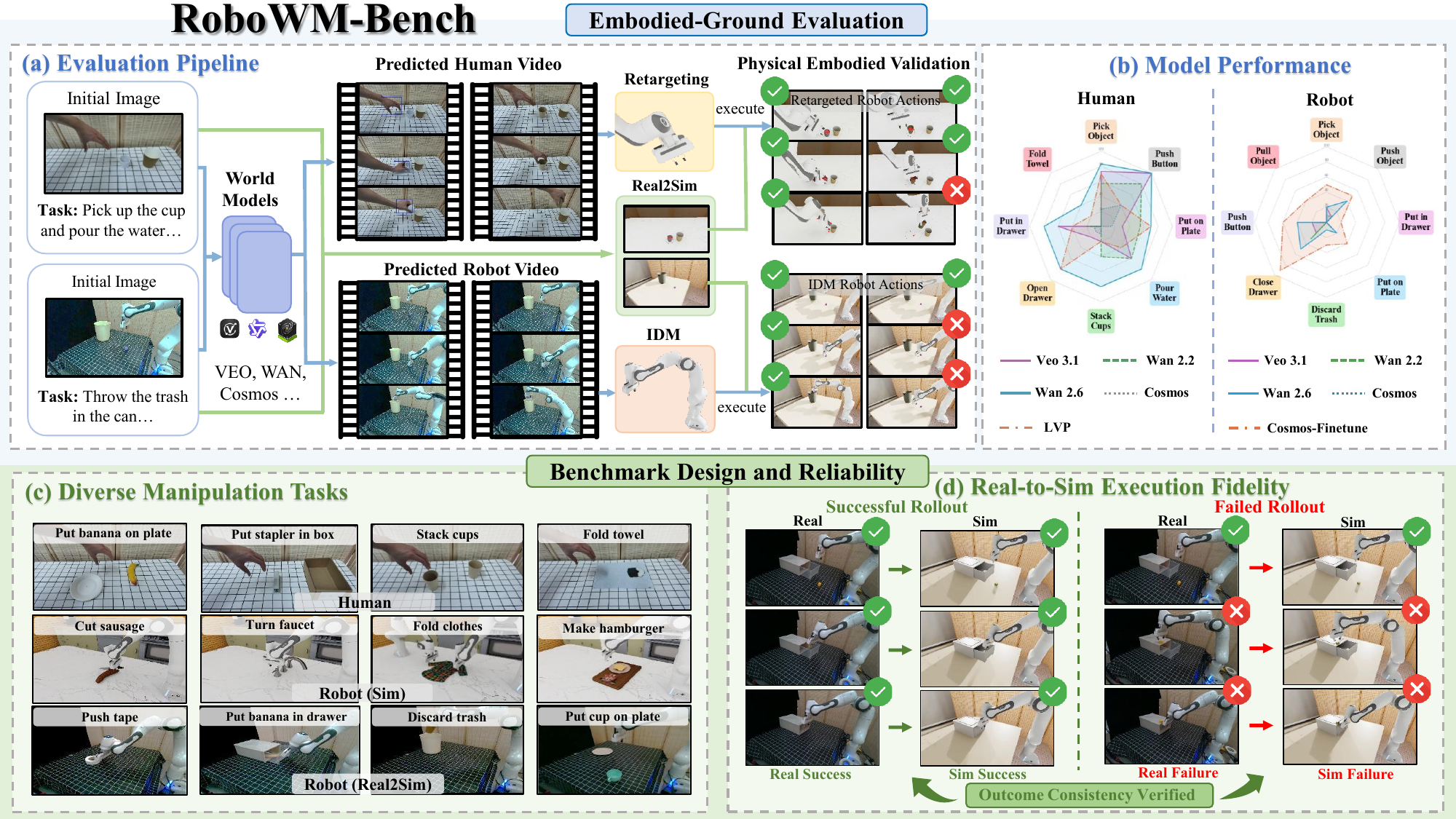}
\vspace{-7mm}
\caption{
\textbf{Overview of RoboWM-Bench.} 
RoboWM-Bench evaluates video world models through embodiment-grounded execution.
(a) Predicted human-hand and robotic manipulation videos are converted into executable actions via retargeting or inverse dynamics model (IDM) and validated in real-to-sim simulation environments.
(b) The benchmark compares execution performance across state-of-the-art video world models for human-hand and robotic manipulation.
(c) RoboWM-Bench covers diverse tasks spanning different object properties, interaction dynamics, temporal horizons, and coordination requirements.
(d) Real-to-sim fidelity is evaluated by replaying identical real-world trajectories in reconstructed simulations and comparing the resulting task outcomes.
} 
\vspace{-2mm}
\label{fig:teaser}
\end{figure*}

Recent advances in large-scale video generation have produced increasingly realistic video world models capable of synthesizing temporally coherent and visually plausible futures~\cite{sora2024worldsimulators,veo3_tech_report_2025,wan2025wan,gao2025seedance}. While these models create new opportunities for robot learning from predicted videos, embodied manipulation requires more than visual fidelity: generated interactions must respect physical constraints and support executable control. Accordingly, recent work has explored robotics-oriented video world models that better capture robot--environment interactions and embodiment-aware dynamics~\cite{ali2025world,chen2025large,chi2025wow,huang1895enerverse,team2025gigaworld,agarwal2025cosmos}. As predicted manipulation videos become a promising source of scalable supervision for robot learning~\cite{jang2025dreamgen,chen2025large,team2025gigabrain}, reliable evaluation is essential for determining whether they are physically grounded and useful for downstream control.

Existing benchmarks for video world models have provided valuable tools for evaluating visual fidelity, semantic consistency, and temporal coherence~\cite{huang2024vbench,feng2024tc,ling2025vmbench,ji2024t2vbench,liu2024evalcrafter,yue2025ewmbench}. More recent efforts further extend evaluation toward physical plausibility, offering useful diagnostics of whether generated videos preserve coherent dynamics~\cite{zheng2025vbench,zhou2025pai,shang2026worldarena}. For robotic manipulation, however, physical plausibility raises a more operational question: can the behaviors depicted in predicted videos be translated into executable actions that accomplish the intended task? While recent work~\cite{fan2026wow} takes an important step toward embodiment-grounded evaluation through real-world robot validation, broad and reproducible assessment remains challenging due to the cost and limited scalability of physical experiments.

To address this gap, we introduce \textbf{RoboWM-Bench}, a manipulation-centric benchmark for evaluating video world models through execution-based validation in high-fidelity simulation environments. RoboWM-Bench covers both human-hand and robotic manipulation scenarios, reflecting complementary settings for learning from human demonstrations and robot-centric control. 
As illustrated in Figure~\ref{fig:teaser}, for each task, a video world model is conditioned on an initial observation and task description to generate a future manipulation video. The predicted behaviors are then converted into executable action sequences through inverse dynamics modeling~\cite{jang2025dreamgen,baker2022video} for robotic videos, or pose tracking and retargeting~\cite{pavlakos2024reconstructing,lepert2503phantom,lepert2025masquerade} for human demonstrations. These actions are executed in high-fidelity real-to-simulation environments, enabling standardized and reproducible validation of physical executability. RoboWM-Bench combines step-level executability checks with final task-level success rates, supporting both fine-grained diagnosis and holistic assessment. Spanning diverse object dynamics, task horizons, and single-arm and bimanual interactions, RoboWM-Bench provides an accessible and scalable protocol for comparing video world models under embodiment constraints.

We conduct extensive experiments with RoboWM-Bench to evaluate state-of-the-art video world models under embodied execution. The results suggest that visual realism does not always translate into physical executability, with execution success tending to decrease as task complexity increases, especially in long-horizon interactions and deformable-object manipulation. Qualitative analysis identifies recurring sources of execution difficulty, including inaccurate contact prediction, unrealistic object deformation, and geometric distortions that can lead to dynamically infeasible actions. Although fine-tuning on manipulation-specific data improves executability, generating physically consistent interactions remains challenging. We further validate the evaluation pipeline through action-extraction accuracy and real-to-sim execution consistency analyses, supporting the reliability of these findings. Together, these results demonstrate the value of RoboWM-Bench as a diagnostic protocol for embodiment-grounded evaluation and highlight opportunities for advancing physically grounded world models for robotic manipulation.

\section{Related Work}
\label{sec:related}


\vspace{-2mm}
\subsection{World Models for Robotics}
\vspace{-2mm}

Recent advances in large-scale video generation have renewed interest in world models as predictive models of physical dynamics~\cite{wu2025hunyuanvideo,team2025klingavatar,hong2022cogvideo,wang2025lavie,hacohen2024ltx,agarwal2025cosmos,ali2025world}. Models such as Sora~\cite{sora2024worldsimulators}, Veo~\cite{veo3_tech_report_2025}, Wan~\cite{wan2025wan}, and Seedance~\cite{gao2025seedance} demonstrate strong visual realism and temporal coherence, suggesting that large-scale video training can capture rich spatiotemporal priors. For robotic manipulation, however, the key question is not only whether generated videos look plausible, but whether they preserve the physical and interaction dynamics needed for executable control. 
This has motivated robotics-oriented world models for video-based robot planning~\cite{assran2025v,chen2025large}, scalable data generation for robot learning~\cite{jang2025dreamgen,team2025gigaworld}, embodied interaction prediction and interactive real-world simulation~\cite{yang2023learning,chi2025wow,huang1895enerverse}, and robot imagination with compositional video planning~\cite{zhou2024robodreamer}.
As video world models are increasingly used as simulators, planners, or data engines for embodied AI, their evaluation must move beyond perceptual quality to assess physical consistency and control feasibility.

\vspace{-2mm}
\subsection{Learning Robotic Actions from Video}
\vspace{-2mm}

Learning robotic control from video has become increasingly important as large-scale human and generated videos provide scalable supervision for robot learning~\cite{radosavovic2023real,bharadhwaj2024gen2act,qin2022dexmv,grauman2022ego4d,xiong2021learning,mccarthy2025towards,team2025evaluating}. 
Prior work has explored several ways to bridge videos and actions. 
Representation-learning methods such as R3M~\cite{nair2022r3m} and VIP~\cite{ma2022vip} pretrain visual encoders on diverse videos to improve downstream policy learning. 
Latent-action and domain-adaptation methods infer action-relevant structure without dense robot action annotations~\cite{ye2024latent,punamiya2025egobridge}, while inverse-dynamics approaches recover executable actions from observed or generated videos~\cite{du2023learning,jang2025dreamgen,tian2024predictive}. 
Human-to-robot transfer methods further retarget in-the-wild human videos to robot policies, as in Phantom~\cite{lepert2503phantom} and Masquerade~\cite{lepert2025masquerade}. 
Recent VLA and world-model-based approaches also use videos as training or planning signals~\cite{kareer2025emergence,jang2025dreamgen,team2025gigabrain,bjorck2025gr00t,luo2024grounding}. 
These advances make video-to-action interfaces increasingly practical; in this work, we use them not as a policy-learning objective, but as evaluation tools for testing whether generated manipulation videos can support successful embodied execution.

\vspace{-2mm}
\subsection{Evaluation and Benchmarks of World Models}
\vspace{-2mm}

Existing benchmarks for video world models primarily assess perceptual realism and generative quality~\cite{feng2024tc,ling2025vmbench}. 
Comprehensive suites such as VBench~\cite{huang2024vbench}, EvalCrafter~\cite{liu2024evalcrafter}, and T2VEval~\cite{ji2024t2vbench} evaluate visual fidelity, temporal consistency, text--video alignment, and motion coherence. 
More recent physical-AI-oriented benchmarks~\cite{li2025worldmodelbench,yue2025ewmbench,shang2026worldarena}, including PAI-Bench~\cite{zhou2025pai} and VBench-2.0~\cite{zheng2025vbench}, further extend evaluation toward physical reasoning, intrinsic faithfulness, and embodied-world-model capabilities. 
These benchmarks provide valuable and reproducible diagnostics, while leaving action-level task completion under embodied constraints less directly measured.

Several robotics-oriented studies have begun to incorporate execution-based validation. 
For example, LVP~\cite{chen2025large} evaluates video-conditioned planning on real-world manipulation tasks, and Wow-wo-val~\cite{fan2026wow} introduces embodied evaluation protocols inspired by Turing-test-style assessment. 
These works represent important steps toward physically grounded evaluation; however, real-robot evaluation typically requires dedicated hardware setups, which can limit scalability for broad model comparisons. 
Complementary to these efforts, RoboWM-Bench provides a manipulation-centric benchmark with standardized task suites and a unified execution-based protocol across simulation-native and real-to-sim reconstructed scenarios. 
This enables scalable assessment of physical executability and action-level consistency in predicted manipulation videos.
\vspace{-1mm}
\section{RoboWM-Bench}
\label{sec:benchmark}

\vspace{-2mm}
\subsection{Benchmark Overview}
\vspace{-2mm}

\begin{figure}[t]
\centering
\includegraphics[width=\linewidth]{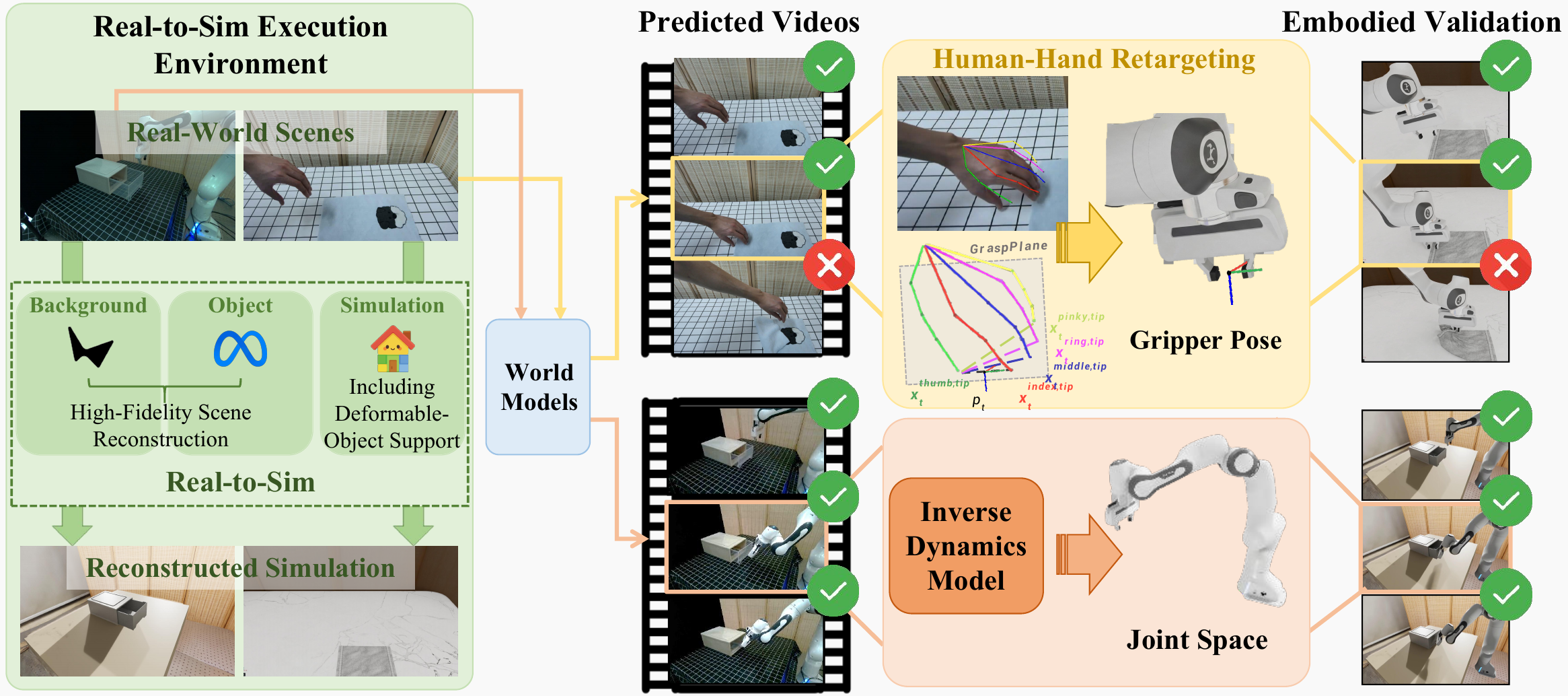}
\vspace{-7mm}
\caption{
\textbf{Pipeline of RoboWM-Bench.} Given an initial scene observation, the corresponding real-world scene is reconstructed in simulation through a real-to-sim pipeline, enabling consistent and reproducible evaluation. Predicted videos are then converted into executable robot actions through two pathways: human-centric retargeting, which estimates 3D hand poses and retargets them to robot end-effector actions, and robot-centric inverse dynamics, which predicts joint-space actions via an inverse dynamics model (IDM). The resulting actions are executed in simulation and evaluated using step-level checkers and final task success rates to measure embodied executability.
} 
\label{fig:framework}
\vspace{-2mm}
\end{figure}

We introduce RoboWM-Bench, a benchmark for evaluating video world models through embodiment-grounded validation, as illustrated in Figure~\ref{fig:framework}. RoboWM-Bench treats physical executability as a measurable criterion by testing whether behaviors inferred from generated videos can complete manipulation tasks when executed. First, RoboWM-Bench provides standardized simulation environments for reproducible evaluation, covering both simulation-native scenarios and real-to-sim reconstructions of real-world scenes (Section~\ref{subsec:real2sim}). Given an initial scene observation and a task description, a video world model predicts a future manipulation video in either a human-hand or robotic setting. The benchmark then provides video-to-action interfaces that convert the generated videos into action sequences, which are executed in the corresponding simulation environments (Section~\ref{subsec:action_extraction}). Finally, RoboWM-Bench defines a diverse manipulation task suite spanning different object properties, interaction regimes, and temporal horizons (Section~\ref{subsec:task}), and evaluates executability through step-level verification and final task-level success (Section~\ref{subsec:evaluation}).

\vspace{-2mm}
\subsection{High-Fidelity Real-to-Sim Framework} \label{subsec:real2sim}
\vspace{-1mm}

To support accessible, reproducible, and scalable evaluation, RoboWM-Bench conducts embodiment-grounded validation in open-source simulation environments. It covers both simulation-native scenarios and real-to-sim reconstructions of real-world scenes, enabling predicted interactions to be evaluated under controlled physical dynamics while reducing dependence on specific physical setups.

To support high-fidelity execution, RoboWM-Bench builds on the LeHome simulation engine~\cite{li2025lehome}, which supports household manipulation scenarios involving rigid, articulated, and deformable objects. We design simulation-native scenarios using LeHome engine, and reconstruct real-world scenes with a modular real-to-sim pipeline covering scene reconstruction, object modeling, and pose calibration. Inspired by recent work~\cite{worldlabs2025marble}, background scenes are reconstructed using 4D Gaussian representations to preserve visual realism and spatial consistency. For interactive objects, rigid geometries are obtained through 3D segmentation and reconstruction~\cite{chen2025sam}, while articulated and deformable object pairs are constructed following~\cite{li2025lehome}. Object poses are estimated using pose estimation models~\cite{wen2024foundationpose,labbe2022megapose}, and camera poses are calibrated with FEEPE~\cite{wu2025foundation}, averaging results across multiple runs for stability. We further validate the fidelity of the reconstructed environments in Section~\ref{subsec:validation}. This modular design preserves the physical structure and spatial configuration of real scenes and supports new evaluation scenarios through either simulation-native assets or reconstructed real-world scenes.

\vspace{-2mm}
\subsection{Embodied Video-to-Action Execution} \label{subsec:action_extraction}
\vspace{-1mm}
To evaluate generated videos through execution, RoboWM-Bench converts predicted manipulation videos into executable action sequences. The benchmark supports both human-hand and robotic videos, covering complementary evaluation settings: human-hand videos reflect interaction patterns that current world models often generate more reliably, while robotic videos are more directly aligned with downstream robot policy learning and control. We therefore use separate video-to-action pathways for the two settings.

\vspace{-2mm}
\subsubsection{Human-Centric Retargeting}
\vspace{-2mm}
Inspired by prior work~\cite{lepert2503phantom, lepert2025masquerade}, we estimate human hand poses from videos and retarget them to robot end-effector actions. We reconstruct 3D hand poses using HaMeR~\cite{pavlakos2024reconstructing} and use the recovered keypoints to derive three control components: end-effector position, orientation, and gripper opening.

The gripper target position is defined as the midpoint between the thumb and index fingertips. For orientation, instead of relying on global finger configurations as in prior work~\cite{lepert2503phantom}, we use contact-relevant geometry: we fit a plane through the thumb and index finger keypoints, project the two fingertips onto this plane, and define the $x$-axis as the line connecting the projections, with the $z$-axis given by the plane normal. This yields more stable end-effector poses while preserving local human--object interaction geometry. For gripper opening, we use the minimum distance between the thumb tip and all other fingertips rather than the thumb--index distance, accounting for cases where the index fingertip is not the primary contact point. We then apply trajectory smoothing and temporal denoising to stabilize the retargeted motion signals~\cite{lepert2503phantom, lepert2025masquerade}.

\subsubsection{Robot-Centric Execution}
For robotic manipulation videos, we recover action sequences using an inverse dynamics model (IDM), following established video-to-action formulations in prior work~\cite{jang2025dreamgen,baker2022video}. We adopt the IDM architecture from~\cite{jang2025dreamgen}, which takes two consecutive image frames as input and predicts the intermediate joint-space action chunk. We train the IDM with a two-stage strategy: large-scale simulation data provides action-labeled supervision for motion pretraining, while a small set of real-world trajectories is used for visual adaptation.

Specifically, we first collect simulation trajectories with a Franka arm in a physics simulator, recording paired RGB observations and joint-space actions. Compared with real-world collection, simulation enables action-labeled trajectories to be generated more efficiently and at higher temporal resolution, providing dense and smooth supervision for inverse-dynamics learning. To keep simulation pretraining scalable and focused on robot motion, we apply background masking, inspired by~\cite{team2025gigaworld}, retaining only the robot arm in simulation videos. We then finetune the IDM on real-world trajectories collected with a physical Franka arm, without background masking. This design leverages scalable simulation supervision while adapting to real observations, enabling reliable action extraction from generated robotic videos for embodiment-grounded evaluation.

\subsection{Manipulation Task Suite with Diverse Complexity} \label{subsec:task}

RoboWM-Bench includes a diverse suite of manipulation tasks designed to evaluate the embodied reasoning and physical consistency of video world models across varying levels of complexity. The tasks span different object properties, interaction dynamics, temporal horizons, and coordination requirements, enabling systematic assessment beyond short-horizon rigid-object manipulation.

The task suite covers several representative interaction regimes. Basic rigid-object tasks, such as object pickup and trash disposal, primarily assess contact precision and spatial reasoning. Articulated-object tasks, including drawer opening and faucet rotation, require models to capture kinematic constraints and structured motion. Deformable-object tasks, such as towel folding, further evaluate whether models can generate physically plausible non-rigid interactions.

RoboWM-Bench also includes more complex manipulation settings that require longer-term reasoning or bimanual coordination. Long-horizon compositional tasks, such as assembling a hamburger, test multi-stage planning and temporal consistency, while bimanual tasks, such as object handover and collaborative towel folding, introduce coordination constraints between two hands. Together, this structured task design supports fine-grained evaluation of how video world models handle increasingly complex embodied interactions.

\subsection{Evaluation of Embodied Executability} \label{subsec:evaluation}

We define \emph{embodied executability} as whether behaviors inferred from generated videos can be executed to complete the intended task. RoboWM-Bench evaluates executability at two levels: step-level verification and task-level success.
For each task, we predefine a set of key action nodes corresponding to semantically meaningful interaction stages, such as contact events (\emph{e.g.}, grasping) or stable end-effector configurations (\emph{e.g.}, lifting). During execution, step-level verification checks whether the predicted behavior satisfies the required interaction and dynamical constraints at each key node. A trajectory is considered task-level successful only if all key nodes pass these checks and the final task objective is achieved. This hierarchical protocol supports fine-grained failure diagnosis while providing a clear and measurable criterion for overall task completion.

\vspace{-1mm}
\section{Experiments}
\label{sec:exp}
\vspace{-2mm}

We evaluate state-of-the-art video world models on RoboWM-Bench across a diverse set of human-hand and robotic manipulation tasks (Section~\ref{subsec:exp_setup}). Our experiments first quantify embodied executability and characterize common failure modes (Section~\ref{subsec:exp_execution}). We then compare RoboWM-Bench with PAI-Bench on the same generated videos to examine the complementary role of execution-based evaluation (Section~\ref{subsec:with_pai_bench}). Finally, we validate the reliability of RoboWM-Bench through action-extraction and simulation-consistency analyses (Section~\ref{subsec:validation}).

\vspace{-1mm}
\subsection{Environment Setup}
\label{subsec:exp_setup}
\vspace{-1mm}
\subsubsection{Tasks and Environments}
\vspace{-1mm}

We evaluate video world models on the RoboWM-Bench task suite, which covers both human-hand and robotic manipulation tasks. In each robotic task setup, a Franka arm operates on a tabletop where objects are initialized with randomized poses, and the scene is observed from a camera viewpoint that remains fixed during evaluation. We consider both simulation-native scenarios and real-world scenes reconstructed through our real-to-sim pipeline, enabling generated robotic behaviors to be evaluated under a unified simulation-based execution protocol.

For human-hand tasks, we use real-world scenes with randomized object poses. Following the input convention of LVP~\cite{chen2025large}, the initial observation includes a visible human hand positioned above the tabletop, ensuring fair comparison across baselines. We focus on real-world scenes because simulating a real human hand within the physics engine would not faithfully capture human manipulation dynamics and would therefore provide limited evaluation value.

All executions are conducted in the LeHome simulation environment~\cite{li2025lehome,isaacsim2023}. To ensure fair and reproducible comparison, each task uses 10 different initial object configurations shared across all models, together with standardized task descriptions, random seeds, and evaluation protocols. We report both task- and step-level success rates.

\begin{table*}[t!]
\caption{
Embodied execution success rates (\%) on RoboWM-Bench for human-hand (top two sections) and robotic manipulation tasks (bottom two sections), reported at both task and step levels.
}	
\label{tab:main_results}	
\centering
{\scriptsize
\setlength{\tabcolsep}{3pt}
\renewcommand{\arraystretch}{0.8}

\begin{tabularx}{\textwidth}{l|*{8}{|>{\centering\arraybackslash}X}}
\toprule

& \multicolumn{8}{c}{\textbf{Human (Task Level)}} \\
\cmidrule(lr){2-9}

\textbf{Method} & Pick Object & Push Button & Put on Plate & Pour Water & Stack Cups & Open Drawer & \scalebox{0.9}[1.0]{Put in Drawer} & Fold Towel \\
\midrule
Cosmos    & 23\% & 40\% & 15\% & 0\%  & 10\% & 10\% & 10\% & 0\% \\
Wan 2.2   & 57\% & 80\% & 55\% & 60\% & 40\% & 0\%  & 20\% & 0\% \\
Veo 3.1   & 73\% & \textbf{100\%} & 30\% & 60\% & 20\% & 20\% & 60\% & 0\% \\
Wan 2.6   & \textbf{83\%} & \textbf{100\%} & \textbf{70\%} & \textbf{80\%} & \textbf{80\%} & \textbf{80\%} & \textbf{80\%} & \textbf{40\%} \\
LVP       & 70\% & 40\% & \textbf{70\%} & 40\% & 20\% & \textbf{80\%} & 40\% & 20\% \\
\midrule
\midrule

& \multicolumn{8}{c}{\textbf{Human (Step Level)}} \\
\cmidrule(lr){2-9}
\textbf{Method} & \multicolumn{3}{c|}{\textbf{Put on Plate}} & \multicolumn{5}{c}{\textbf{Put in Drawer}}  \\
& contact & lift & place & contact & lift & above drawer & in drawer & close drawer \\
\midrule
Cosmos  & 90\% & 20\% & 15\% & 80\% & 20\% & 20\% & 20\% & 10\% \\
Wan 2.2 & \textbf{100\%} & 60\% & 55\% & \textbf{100\%} & 60\% & 60\% & 40\% & 20\% \\
Veo 3.1 & \textbf{100\%} & 70\% & 30\% & \textbf{100\%} & 70\% & 70\% & 60\% & 60\% \\
Wan 2.6 & \textbf{100\%} & \textbf{75\%} & \textbf{70\%} & \textbf{100\%} & \textbf{80\%} & \textbf{80\%} & \textbf{80\%} & \textbf{80\%} \\
LVP     & \textbf{100\%} & \textbf{75\%} & \textbf{70\%} & \textbf{100\%} & 70\% & 60\% & 50\% & 40\% \\

\midrule
\midrule

& \multicolumn{8}{c}{\textbf{Robot (Task Level)}} \\
\cmidrule(lr){2-9}
\textbf{Method} & Close Drawer & Pick Object & Push Object & Push Button & Put on Plate & \scalebox{0.9}[1.0]{Discard Trash} & Pull Object & \scalebox{0.9}[1.0]{Put in Drawer} \\
\midrule
Cosmos    & 0\%  & 10\% & 10\% & 10\% & 10\% & 0\%  & 0\% & 0\% \\
Wan 2.2   & 30\% & 10\% & 0\%  & 0\%  & 0\%  & 0\%  & 0\% & 0\% \\
Veo 3.1      & 20\% & 20\% & 10\% & 20\% & 10\%  & 0\%  & 0\% & 0\% \\
Wan 2.6   & 50\% & 20\% & 40\% & 40\% & 20\% & 10\% & 0\% & 0\% \\
Cosmos-FT & \textbf{90\%} & \textbf{50\%} & \textbf{50\%} & \textbf{60\%} & \textbf{40\%} & \textbf{30\%} & \textbf{40\%} & \textbf{20\%} \\

\midrule
\midrule

& \multicolumn{8}{c}{\textbf{Robot (Step Level)}} \\
\cmidrule(lr){2-9}
\textbf{Method} & \multicolumn{3}{c|}{\textbf{Put on Plate}} & \multicolumn{5}{c}{\textbf{Put in Drawer}} \\
& contact & lift & place & contact & lift & above drawer & in drawer & close drawer \\
\midrule
Cosmos    & 30\% & 10\% & 10\% & 10\% & 0\% & 0\% & 0\% & 0\% \\
Wan 2.2   & 20\% & 0\% & 0\% & 0\% & 0\% & 0\% & 0\% & 0\% \\
Veo 3.1      & 40\% & 10\% & 10\% & 30\% & 0\% & 0\% & 0\% & 0\% \\
Wan 2.6   & 40\% & 20\% & 20\% & 30\% & 0\% & 0\% & 0\% & 0\% \\
Cosmos-FT & \textbf{60\%} & \textbf{40\%} & \textbf{40\%} & \textbf{60\%} & \textbf{20\%} & \textbf{20\%} & \textbf{20\%} & \textbf{20\%} \\

\bottomrule
\end{tabularx}
}
\vspace{-2mm}
\end{table*}

\vspace{-2mm}
\subsubsection{Baselines}
\vspace{-2mm}

We evaluate state-of-the-art video world models covering both general-purpose and interaction-oriented generation. For general-purpose video generation, we include closed-source models, Veo3.1~\cite{veo3_tech_report_2025} and Wan2.6~\cite{wan2025wan}, as well as open-source models, Wan2.2~\cite{wan2025wan} and Cosmos-Predict2.5~\cite{ali2025world}. For interaction-oriented generation, we include LVP~\cite{chen2025large}, which is specifically trained to capture complex human interactive behaviors. To further investigate the potential of video world models for embodied manipulation, we fine-tune Cosmos-Predict2.5 on our collected real-world manipulation dataset with 50 trajectories per task and denote the resulting variant as Cosmos-Finetune.

\vspace{-1mm}
\subsection{Embodied Executability of Video World Models}
\label{subsec:exp_execution}
\vspace{-2mm}

Table~\ref{tab:main_results} reports task- and step-level execution success rates. Results on simulation-native robotic tasks are provided in the supplementary due to space limitations (Section~\ref{sec:robot_sim}). The experimental results reveal several consistent trends, which we analyze below.

First, \textbf{human-hand videos achieve higher execution success than robotic manipulation videos.} This gap likely reflects biases in large-scale video pretraining, where human interactions are far more prevalent than robotic data. Generated human hands also tend to preserve more stable geometry during interaction, whereas robotic manipulators are more likely to exhibit structural distortions that lead to execution failures.

Second, \textbf{execution success decreases as task complexity increases.} Short-horizon tasks such as \textit{Push Button} are generally easier, while longer-horizon tasks such as \textit{Put in Drawer} suffer from error accumulation across multiple interaction stages. Among human-hand tasks, \textit{Fold Towel} remains particularly challenging, reflecting the added complexity of deformable-object interactions.

Third, \textbf{manipulation-specific fine-tuning improves robotic executability.} Cosmos-Finetune substantially outperforms its pretrained counterpart, indicating that even limited task-specific data, with 50 trajectories per task, can improve the generation of executable robotic behaviors. Fine-tuning reduces robot-geometry artifacts and improves joint articulation, although errors in 3D object localization and grasp stability remain common failure factors.

\begin{figure*}[t]
\centering
    \includegraphics[width=\linewidth]{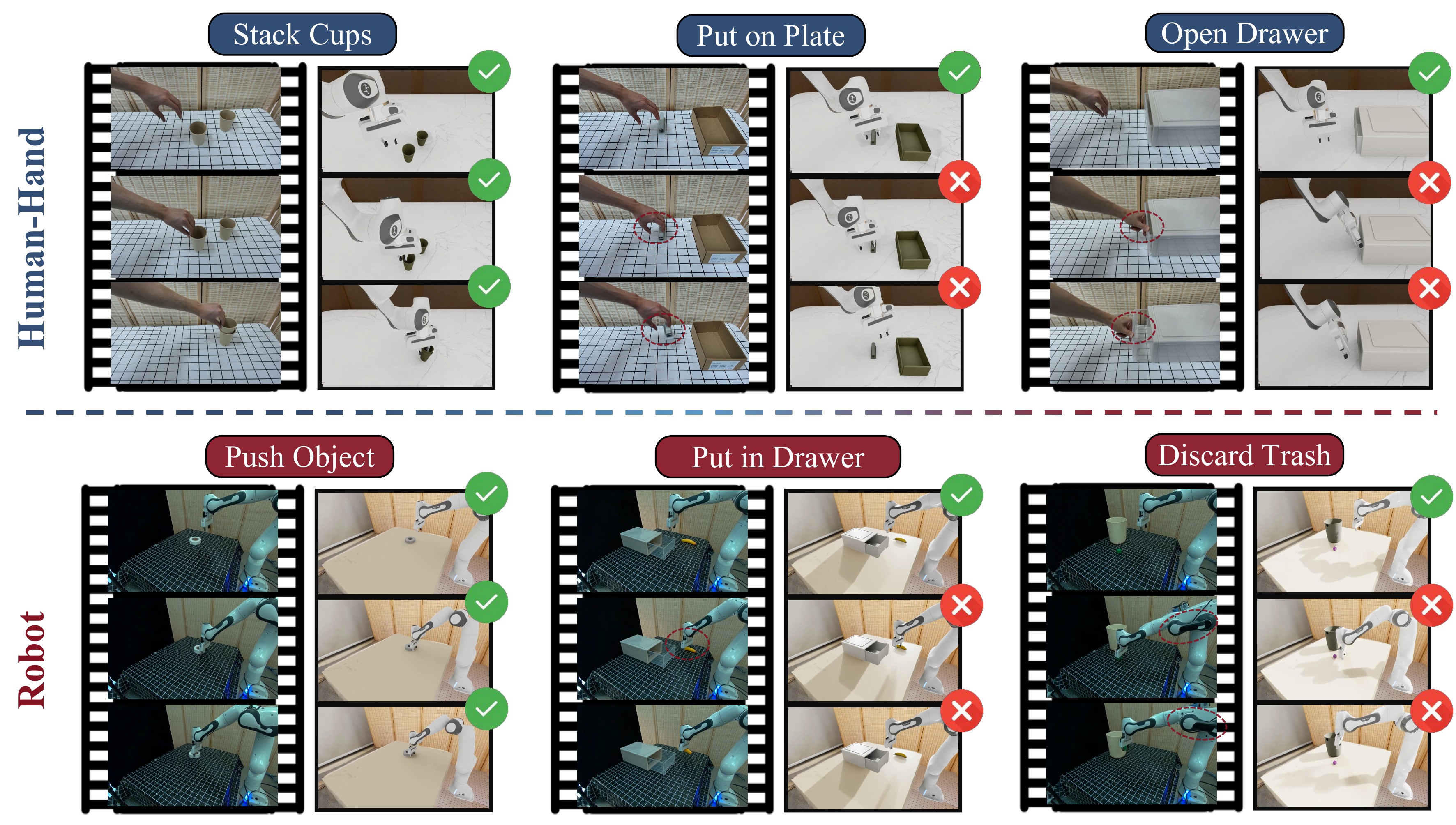}
\vspace{-7mm}
\caption{
\textbf{Qualitative execution results on RoboWM-Bench.} 
For each task, predicted videos (left) are converted into robot actions and executed in simulation (right).
} 
\vspace{-2mm}
\label{fig:results}
\end{figure*}

Figure~\ref{fig:results} presents representative qualitative results from Wan2.6, the strongest overall model in our evaluation. Its success and failure cases therefore provide useful insight into the remaining gap between visual plausibility and embodied executability. These examples show that generated interactions can appear plausible in video while still failing under physical execution. In \emph{Put on Plate}, for instance, the predicted video shows the fingers merely touching the object without forming a stable grasp, yet the object is lifted. Such interactions are physically implausible and fail during execution. In \emph{Open Drawer}, the predicted motion resembles closing the drawer and does not establish a proper grasp, which is reflected by the simulated execution outcome. 
For robotic manipulation videos, failures can also arise from geometric distortions of the generated robot arm. The end effector may appear to reach the target in the video, while the recovered joint configuration yields a mismatched end-effector pose under the robot's true forward kinematics.

\vspace{-1mm}
\subsection{Perceptual Plausibility vs. Embodied Executability}
\label{subsec:with_pai_bench}
\vspace{-2mm}

\begin{figure*}[t]
\centering
\includegraphics[width=\linewidth]{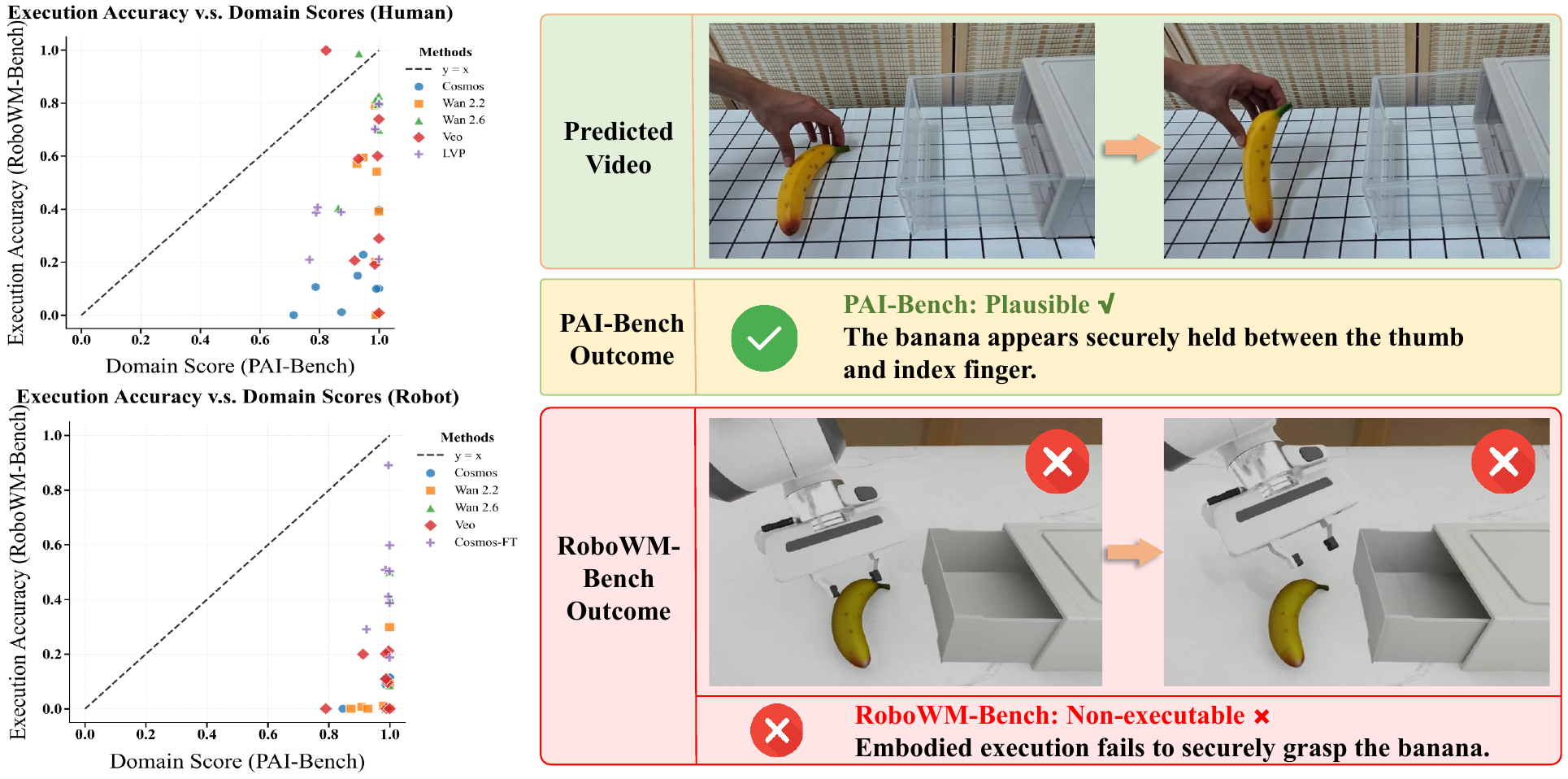}
\vspace{-7mm}
\caption{
Comparison between PAI-Bench and RoboWM-Bench. 
} 
\label{fig:with_pai_bench}
\vspace{1mm}
\end{figure*}

We compare execution accuracy in RoboWM-Bench with the domain scores in PAI-Bench, a commonly used metric for evaluating the perceptual plausibility of generated videos. As shown in Figure~\ref{fig:with_pai_bench}, the same predicted videos obtain near-saturated scores on PAI-Bench across different world models, whereas RoboWM-Bench evaluates whether the predicted behaviors are physically executable, resulting in more discriminative outcomes. This discrepancy arises because some actions may appear visually plausible (Figure~\ref{fig:results}) yet remain physically infeasible, an issue that perceptual domain scores may not capture but becomes evident under embodied execution. These results highlight the complementary role of RoboWM-Bench and show that embodiment-grounded evaluation provides a more direct measure of physical executability.
Further PAI-Bench analysis and domain-score implementation details are provided in Appendix~\ref{sec:pai_quality_score} and Appendix~\ref{sec:pai_domain_implementation}, respectively.

\vspace{-1mm}
\subsection{Reliability Analysis of RoboWM-Bench}
\label{subsec:validation}
\vspace{-2mm}

\begin{figure}[t]
    \centering
    \begin{minipage}[t]{0.56\linewidth}
        \vspace{0pt}
        \centering
        \includegraphics[width=\linewidth]{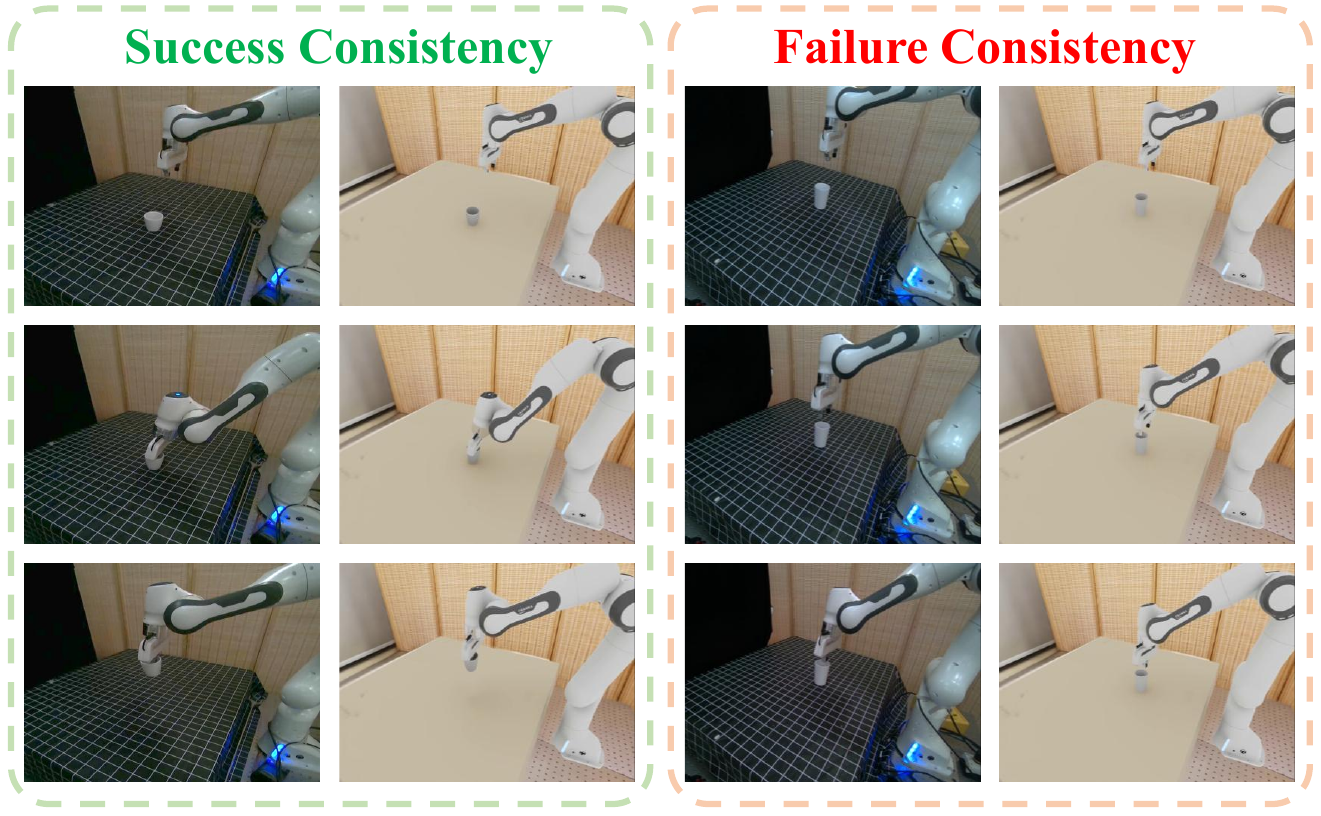}
    \end{minipage}
    \hfill
    \begin{minipage}[t]{0.42\linewidth}
        \vspace{7pt}
        \centering
        \scriptsize
        \renewcommand{\arraystretch}{1.4}
        \begin{tabular}{lccc}
            \toprule
            Task & \makecell{Success\\Consistency} & \makecell{Failure\\Consistency} \\
            \midrule
            Pick Object & 10/10 & 10/10  \\
            Pull Object & 10/10 & 10/10  \\
            Push Object & 10/10 & 10/10  \\
            Put on Plate & 10/10 & 10/10  \\
            Discard Trash & 10/10 & 10/10  \\
            Close Drawer & 10/10 & 10/10  \\
            Put in Drawer & 10/10 & 10/10  \\
            \midrule
            Average & 100\% & 100\% \\
            \bottomrule
        \end{tabular}
    \end{minipage}
    \vspace{-2mm}
    \caption{
    \textbf{Real-to-sim consistency evaluation.} Left: qualitative examples of replaying identical real-world trajectories in reconstructed simulation environments. Right: outcome agreement between real-world and reconstructed simulation executions.
    }
    \vspace{-1mm}
    \label{fig:real-sim-visu}
\end{figure}

To examine the reliability of RoboWM-Bench, we validate two key interfaces in the evaluation pipeline: real-to-sim execution consistency and video-to-action replay success. 
The former tests whether reconstructed simulation environments preserve real-world task outcomes, while the latter evaluates whether retargeting or inverse dynamics can recover executable actions from videos.

\vspace{-1mm}
\subsubsection{Real-to-Sim Execution Consistency}
\vspace{-1mm}

We first assess whether reconstructed simulation environments preserve real-world execution outcomes. 
Specifically, we execute real-world robotic trajectories in the corresponding reconstructed simulation environments and compare task outcomes across domains. 
We use robotic trajectories because they provide the same low-level action sequences in both real and simulated scenes, keeping this test focused on reconstruction fidelity; video-to-action conversion, including human-to-robot retargeting, is evaluated separately in the next subsection.

For each task, we collect 10 successful and 10 failed real-world trajectories and execute the same actions in the reconstructed simulation environments. 
Figure~\ref{fig:real-sim-visu} shows that simulation consistently reproduces both success and failure outcomes. 
This consistency suggests that the reconstructed environments preserve the task-relevant geometry and interaction constraints needed for execution-based evaluation.
Real-world and simulated videos are provided in the supplementary material.

\vspace{-1mm}
\subsubsection{Video-to-Action Replay Success}
\vspace{-1mm}

\begin{table*}[t!]
\caption{
Accuracy of action extraction methods.
}	

\label{tab:extraction_accu}
{\scriptsize
\setlength{\tabcolsep}{3pt}
\renewcommand{\arraystretch}{0.8}

\begin{tabularx}{\textwidth}{l|*{8}{|>{\centering\arraybackslash}X}}
\toprule

& \multicolumn{8}{c}{\textbf{Human}} \\
\cmidrule(lr){2-9}
\textbf{Method} & Pick Object & Stack Cups & Pour Water & Open Drawer & Fold Towel & Put on Plate & \scalebox{0.9}[1.0]{Put in Drawer} & Average \\
\midrule
Retargeting   & \textbf{100\%} & \textbf{90\%} & \textbf{90\%} & \textbf{100\%} & \textbf{100\%} & \textbf{100\%} & \textbf{100\%} & \textbf{97.1\%} \\

\midrule
\midrule

& \multicolumn{8}{c}{\textbf{Robot}} \\
\cmidrule(lr){2-9}
\textbf{Method} & Pick Object & Pull Object & Push Object & \scalebox{0.9}[1.0]{Discard Trash} & Close Drawer & Put on Plate & \scalebox{0.9}[1.0]{Put in Drawer} & Average \\
\midrule
IDM$_{\text{Real}}$   & 70\% & 70\% & 80\% & 70\% & 90\% & 70\% & 50\% & 71.4\% \\
IDM$_{\text{Sim+Real}}$    & \textbf{100\%} & \textbf{90\%} & \textbf{100\%}  & \textbf{90\%}  & \textbf{100\%} & \textbf{100\%} & \textbf{90\%} &  \textbf{95.7\%} \\

\bottomrule
\end{tabularx}
}
\small
\vspace{-1mm}
\end{table*}

We next evaluate whether the video-to-action modules can recover executable actions from videos. 
We process videos from successful real-world manipulation trajectories using our action extraction pipeline, and execute the extracted actions in the reconstructed simulation environments to test whether they reproduce the original outcomes.

As shown in Table~\ref{tab:extraction_accu}, the human-hand pose tracking and retargeting pipeline achieves near-perfect replay success. 
The remaining failures in \textit{Stack Cups} and \textit{Pour Water} occur in deliberately contact-sensitive cases involving cylindrical objects, where small discrepancies between human fingertip contacts and retargeted robot gripper contacts can destabilize grasps. 
These cases provide conservative stress tests for retargeting accuracy rather than typical failures observed across most generated videos.

For robotic videos, we compare IDM$_{\text{Real}}$, trained on 50 real-world trajectories per task, with IDM$_{\text{Sim+Real}}$, which uses large-scale simulation pretraining followed by finetuning on the same real-world trajectories. 
IDM$_{\text{Sim+Real}}$ achieves substantially higher replay success, indicating that scalable action-labeled simulation data provides useful motion priors and improves inverse-dynamics prediction. 
The remaining failures mainly arise in near-boundary contact cases, such as intentionally shallow grasps, where minor prediction errors can compromise grasp stability.

\section{Conclusion}
\label{sec:conclusion}
\vspace{-1mm}

We introduced RoboWM-Bench, a manipulation-centric benchmark for evaluating video world models through embodiment-grounded execution. RoboWM-Bench converts predicted human-hand and robotic manipulation videos into executable actions and validates them in standardized simulation and real-to-sim environments using step-level and task-level execution metrics. Covering diverse object dynamics, task horizons, and coordination requirements, the benchmark provides a scalable and reproducible protocol that complements perceptual video evaluation with execution-based assessment. Experiments with state-of-the-art video world models show that embodied execution reveals additional challenges in complex manipulation settings, particularly those requiring precise contact, deformable-object reasoning, or long-horizon consistency. Overall, RoboWM-Bench offers an execution-grounded diagnostic framework for advancing physically consistent world models for robotic manipulation.

\bibliographystyle{splncs04}
\bibliography{main}

@misc{sora2024worldsimulators,
  title = {Video Generation Models as World Simulators},
  author = {Brooks, Tim and Peebles, William and Holmes, Connor and DePue, Will and Guo, Yufei and Jing, Li and Schnurr, David and Taylor, Joe and Luhman, Troy and Ng, Clarence and Wang, Ricky and Ramesh, Aditya and others},
  year = {2024},
  howpublished = {\url{https://openai.com/research/video-generation-models-as-world-simulators}},
  note = {OpenAI Technical Report},
}

@article{gao2025seedance,
  title={Seedance 1.0: Exploring the boundaries of video generation models},
  author={Gao, Yu and Guo, Haoyuan and Hoang, Tuyen and Huang, Weilin and Jiang, Lu and Kong, Fangyuan and Li, Huixia and Li, Jiashi and Li, Liang and Li, Xiaojie and others},
  journal={arXiv preprint arXiv:2506.09113},
  year={2025}
}

@techreport{veo3_tech_report_2025,
  title        = {Veo: a Text-to-Video Generation System (Veo-3 Technical Report)},
  author       = {{Google DeepMind}},
  institution  = {Google DeepMind},
  year         = {2025},
  number       = {Veo-3-Tech-Report},
  url          = {https://storage.googleapis.com/deepmind-media/veo/Veo-3-Tech-Report.pdf},
  note         = {Technical Report},
}

@article{wan2025wan,
  title={Wan: Open and advanced large-scale video generative models},
  author={Wan, Team and Wang, Ang and Ai, Baole and Wen, Bin and Mao, Chaojie and Xie, Chen-Wei and Chen, Di and Yu, Feiwu and Zhao, Haiming and Yang, Jianxiao and others},
  journal={arXiv preprint arXiv:2503.20314},
  year={2025}
}

@article{ali2025world,
  title={World simulation with video foundation models for physical ai},
  author={Ali, Arslan and Bai, Junjie and Bala, Maciej and Balaji, Yogesh and Blakeman, Aaron and Cai, Tiffany and Cao, Jiaxin and Cao, Tianshi and Cha, Elizabeth and Chao, Yu-Wei and others},
  journal={arXiv preprint arXiv:2511.00062},
  year={2025}
}

@article{chen2025large,
  title={Large video planner enables generalizable robot control},
  author={Chen, Boyuan and Zhang, Tianyuan and Geng, Haoran and Song, Kiwhan and Zhang, Caiyi and Li, Peihao and Freeman, William T and Malik, Jitendra and Abbeel, Pieter and Tedrake, Russ and others},
  journal={arXiv preprint arXiv:2512.15840},
  year={2025}
}

@article{chi2025wow,
  title={Wow: Towards a world omniscient world model through embodied interaction},
  author={Chi, Xiaowei and Jia, Peidong and Fan, Chun-Kai and Ju, Xiaozhu and Mi, Weishi and Zhang, Kevin and Qin, Zhiyuan and Tian, Wanxin and Ge, Kuangzhi and Li, Hao and others},
  journal={arXiv preprint arXiv:2509.22642},
  year={2025}
}

@article{huang1895enerverse,
  title={Enerverse: Envisioning embodied future space for robotics manipulation (2025)},
  author={Huang, Siyuan and Chen, Liliang and Zhou, Pengfei and Chen, Shengcong and Jiang, Zhengkai and Hu, Yue and Liao, Y and Gao, P and Li, H and Yao, M and others},
  journal={arXiv preprint arXiv:2501.01895},
  year={2025}
}

@article{team2025gigaworld,
  title={Gigaworld-0: World models as data engine to empower embodied ai},
  author={Team, GigaWorld and Ye, Angen and Wang, Boyuan and Ni, Chaojun and Huang, Guan and Zhao, Guosheng and Li, Haoyun and Zhu, Jiagang and Li, Kerui and Xu, Mengyuan and others},
  journal={arXiv preprint arXiv:2511.19861},
  year={2025}
}

@article{jang2025dreamgen,
  title={Dreamgen: Unlocking generalization in robot learning through video world models},
  author={Jang, Joel and Ye, Seonghyeon and Lin, Zongyu and Xiang, Jiannan and Bjorck, Johan and Fang, Yu and Hu, Fengyuan and Huang, Spencer and Kundalia, Kaushil and Lin, Yen-Chen and others},
  journal={arXiv preprint arXiv:2505.12705},
  year={2025}
}

@article{team2025gigabrain,
  title={Gigabrain-0: A world model-powered vision-language-action model},
  author={Team, GigaBrain and Ye, Angen and Wang, Boyuan and Ni, Chaojun and Huang, Guan and Zhao, Guosheng and Li, Haoyun and Li, Jie and Zhu, Jiagang and Feng, Lv and others},
  journal={arXiv preprint arXiv:2510.19430},
  year={2025}
}

@inproceedings{huang2024vbench,
  title={Vbench: Comprehensive benchmark suite for video generative models},
  author={Huang, Ziqi and He, Yinan and Yu, Jiashuo and Zhang, Fan and Si, Chenyang and Jiang, Yuming and Zhang, Yuanhan and Wu, Tianxing and Jin, Qingyang and Chanpaisit, Nattapol and others},
  booktitle={Proceedings of the IEEE/CVF Conference on Computer Vision and Pattern Recognition},
  pages={21807--21818},
  year={2024}
}

@article{zheng2025vbench,
  title={Vbench-2.0: Advancing video generation benchmark suite for intrinsic faithfulness},
  author={Zheng, Dian and Huang, Ziqi and Liu, Hongbo and Zou, Kai and He, Yinan and Zhang, Fan and Gu, Lulu and Zhang, Yuanhan and He, Jingwen and Zheng, Wei-Shi and others},
  journal={arXiv preprint arXiv:2503.21755},
  year={2025}
}

@article{feng2024tc,
  title={Tc-bench: Benchmarking temporal compositionality in text-to-video and image-to-video generation},
  author={Feng, Weixi and Li, Jiachen and Saxon, Michael and Fu, Tsu-jui and Chen, Wenhu and Wang, William Yang},
  journal={arXiv preprint arXiv:2406.08656},
  year={2024}
}

@inproceedings{ling2025vmbench,
  title={Vmbench: A benchmark for perception-aligned video motion generation},
  author={Ling, Xinran and Zhu, Chen and Wu, Meiqi and Li, Hangyu and Feng, Xiaokun and Yang, Cundian and Hao, Aiming and Zhu, Jiashu and Wu, Jiahong and Chu, Xiangxiang},
  booktitle={Proceedings of the IEEE/CVF International Conference on Computer Vision},
  pages={13087--13098},
  year={2025}
}

@inproceedings{ji2024t2vbench,
  title={T2vbench: Benchmarking temporal dynamics for text-to-video generation},
  author={Ji, Pengliang and Xiao, Chuyang and Tai, Huilin and Huo, Mingxiao},
  booktitle={Proceedings of the IEEE/CVF Conference on Computer Vision and Pattern Recognition},
  pages={5325--5335},
  year={2024}
}

@inproceedings{liu2024evalcrafter,
  title={Evalcrafter: Benchmarking and evaluating large video generation models},
  author={Liu, Yaofang and Cun, Xiaodong and Liu, Xuebo and Wang, Xintao and Zhang, Yong and Chen, Haoxin and Liu, Yang and Zeng, Tieyong and Chan, Raymond and Shan, Ying},
  booktitle={Proceedings of the IEEE/CVF conference on computer vision and pattern recognition},
  pages={22139--22149},
  year={2024}
}

@article{yue2025ewmbench,
  title={Ewmbench: Evaluating scene, motion, and semantic quality in embodied world models},
  author={Yue, Hu and Huang, Siyuan and Liao, Yue and Chen, Shengcong and Zhou, Pengfei and Chen, Liliang and Yao, Maoqing and Ren, Guanghui},
  journal={arXiv preprint arXiv:2505.09694},
  year={2025}
}

@article{zhou2025pai,
  title={PAI-Bench: A Comprehensive Benchmark For Physical AI},
  author={Zhou, Fengzhe and Huang, Jiannan and Li, Jialuo and Ramanan, Deva and Shi, Humphrey},
  journal={arXiv preprint arXiv:2512.01989},
  year={2025}
}

@article{fan2026wow,
  title={Wow, wo, val! A Comprehensive Embodied World Model Evaluation Turing Test},
  author={Fan, Chun-Kai and Chi, Xiaowei and Ju, Xiaozhu and Li, Hao and Bao, Yong and Wang, Yu-Kai and Chen, Lizhang and Jiang, Zhiyuan and Ge, Kuangzhi and Li, Ying and others},
  journal={arXiv preprint arXiv:2601.04137},
  year={2026}
}

@article{baker2022video,
  title={Video pretraining (vpt): Learning to act by watching unlabeled online videos},
  author={Baker, Bowen and Akkaya, Ilge and Zhokov, Peter and Huizinga, Joost and Tang, Jie and Ecoffet, Adrien and Houghton, Brandon and Sampedro, Raul and Clune, Jeff},
  journal={Advances in Neural Information Processing Systems},
  volume={35},
  pages={24639--24654},
  year={2022}
}

@inproceedings{pavlakos2024reconstructing,
  title={Reconstructing hands in 3d with transformers},
  author={Pavlakos, Georgios and Shan, Dandan and Radosavovic, Ilija and Kanazawa, Angjoo and Fouhey, David and Malik, Jitendra},
  booktitle={Proceedings of the IEEE/CVF Conference on Computer Vision and Pattern Recognition},
  pages={9826--9836},
  year={2024}
}

@article{lepert2503phantom,
  title={Phantom: Training robots without robots using only human videos},
  author={Lepert, Marion and Fang, Jiaying and Bohg, Jeannette},
  journal={URL https://arxiv. org/abs/2503.00779},
  volume={2},
  year={2025}
}

@article{lepert2025masquerade,
  title={Masquerade: Learning from in-the-wild human videos using data-editing},
  author={Lepert, Marion and Fang, Jiaying and Bohg, Jeannette},
  journal={arXiv preprint arXiv:2508.09976},
  year={2025}
}

@misc{worldlabs2025marble,
    title={{Marble: A Multimodal World Model}},
    author={{World Labs}},
    year={2025},
    url={https://www.worldlabs.ai/blog/marble-world-model},
    note={Accessed: 2026-02},
}

@article{chen2025sam,
  title={Sam 3d: 3dfy anything in images},
  author={Chen, Xingyu and Chu, Fu-Jen and Gleize, Pierre and Liang, Kevin J and Sax, Alexander and Tang, Hao and Wang, Weiyao and Guo, Michelle and Hardin, Thibaut and Li, Xiang and others},
  journal={arXiv preprint arXiv:2511.16624},
  year={2025}
}

@article{zhou2024robodreamer,
  title={Robodreamer: Learning compositional world models for robot imagination},
  author={Zhou, Siyuan and Du, Yilun and Chen, Jiaben and Li, Yandong and Yeung, Dit-Yan and Gan, Chuang},
  journal={arXiv preprint arXiv:2404.12377},
  year={2024}
}

@article{assran2025v,
  title={V-jepa 2: Self-supervised video models enable understanding, prediction and planning},
  author={Assran, Mido and Bardes, Adrien and Fan, David and Garrido, Quentin and Howes, Russell and Muckley, Matthew and Rizvi, Ammar and Roberts, Claire and Sinha, Koustuv and Zholus, Artem and others},
  journal={arXiv preprint arXiv:2506.09985},
  year={2025}
}

@inproceedings{li2025lehome,
  title={LeHome: A Simulation Environment for Deformable Object Manipulation in Household Scenarios},
  author={Li, Zeyi and Yang, Yushi and Xie, Shawn and Xu, Kyle and Chen, Tianxing and Wang, Yuran and Shen, Zhenhao and Shen, Yan and Chen, Yue and Li, Wenjun and Zheng, Yukun and Zhang, Chaorui and Lin, Siyi and Teng, Fei and Yang, Hongjun and Chen, Ming and Xie, Steve and Wu, Ruihai},
  booktitle={IEEE International Conference on Robotics and Automation (ICRA)},
  year={2026}
}

@inproceedings{wen2024foundationpose,
  title={Foundationpose: Unified 6d pose estimation and tracking of novel objects},
  author={Wen, Bowen and Yang, Wei and Kautz, Jan and Birchfield, Stan},
  booktitle={Proceedings of the IEEE/CVF conference on computer vision and pattern recognition},
  pages={17868--17879},
  year={2024}
}

@article{nair2022r3m,
  title={R3m: A universal visual representation for robot manipulation},
  author={Nair, Suraj and Rajeswaran, Aravind and Kumar, Vikash and Finn, Chelsea and Gupta, Abhinav},
  journal={arXiv preprint arXiv:2203.12601},
  year={2022}
}

@article{ma2022vip,
  title={Vip: Towards universal visual reward and representation via value-implicit pre-training},
  author={Ma, Yecheng Jason and Sodhani, Shagun and Jayaraman, Dinesh and Bastani, Osbert and Kumar, Vikash and Zhang, Amy},
  journal={arXiv preprint arXiv:2210.00030},
  year={2022}
}

@article{ye2024latent,
  title={Latent action pretraining from videos},
  author={Ye, Seonghyeon and Jang, Joel and Jeon, Byeongguk and Joo, Sejune and Yang, Jianwei and Peng, Baolin and Mandlekar, Ajay and Tan, Reuben and Chao, Yu-Wei and Lin, Bill Yuchen and others},
  journal={arXiv preprint arXiv:2410.11758},
  year={2024}
}

@inproceedings{punamiya2025egobridge,
  title={Egobridge: Domain adaptation for generalizable imitation from egocentric human data},
  author={Punamiya, Ryan and Patel, Dhruv and Aphiwetsa, Patcharapong and Kuppili, Pranav and Zhu, Lawrence Y and Kareer, Simar and Hoffman, Judy and Xu, Danfei},
  booktitle={Human to Robot: Workshop on Sensorizing, Modeling, and Learning from Humans},
  year={2025}
}

@article{kareer2025emergence,
  title={Emergence of Human to Robot Transfer in Vision-Language-Action Models},
  author={Kareer, Simar and Pertsch, Karl and Darpinian, James and Hoffman, Judy and Xu, Danfei and Levine, Sergey and Finn, Chelsea and Nair, Suraj},
  journal={arXiv preprint arXiv:2512.22414},
  year={2025}
}

@inproceedings{radosavovic2023real,
  title={Real-world robot learning with masked visual pre-training},
  author={Radosavovic, Ilija and Xiao, Tete and James, Stephen and Abbeel, Pieter and Malik, Jitendra and Darrell, Trevor},
  booktitle={Conference on Robot Learning},
  pages={416--426},
  year={2023},
  organization={PMLR}
}

@article{bharadhwaj2024gen2act,
  title={Gen2act: Human video generation in novel scenarios enables generalizable robot manipulation},
  author={Bharadhwaj, Homanga and Dwibedi, Debidatta and Gupta, Abhinav and Tulsiani, Shubham and Doersch, Carl and Xiao, Ted and Shah, Dhruv and Xia, Fei and Sadigh, Dorsa and Kirmani, Sean},
  journal={arXiv preprint arXiv:2409.16283},
  year={2024}
}

@inproceedings{qin2022dexmv,
  title={Dexmv: Imitation learning for dexterous manipulation from human videos},
  author={Qin, Yuzhe and Wu, Yueh-Hua and Liu, Shaowei and Jiang, Hanwen and Yang, Ruihan and Fu, Yang and Wang, Xiaolong},
  booktitle={European Conference on Computer Vision},
  pages={570--587},
  year={2022},
  organization={Springer}
}

@inproceedings{grauman2022ego4d,
  title={Ego4d: Around the world in 3,000 hours of egocentric video},
  author={Grauman, Kristen and Westbury, Andrew and Byrne, Eugene and Chavis, Zachary and Furnari, Antonino and Girdhar, Rohit and Hamburger, Jackson and Jiang, Hao and Liu, Miao and Liu, Xingyu and others},
  booktitle={Proceedings of the IEEE/CVF conference on computer vision and pattern recognition},
  pages={18995--19012},
  year={2022}
}

@article{li2025worldmodelbench,
  title={Worldmodelbench: Judging video generation models as world models},
  author={Li, Dacheng and Fang, Yunhao and Chen, Yukang and Yang, Shuo and Cao, Shiyi and Wong, Justin and Luo, Michael and Wang, Xiaolong and Yin, Hongxu and Gonzalez, Joseph E and others},
  journal={arXiv preprint arXiv:2502.20694},
  year={2025}
}

@inproceedings{xiong2021learning,
  title={Learning by watching: Physical imitation of manipulation skills from human videos},
  author={Xiong, Haoyu and Li, Quanzhou and Chen, Yun-Chun and Bharadhwaj, Homanga and Sinha, Samarth and Garg, Animesh},
  booktitle={2021 IEEE/RSJ international conference on intelligent robots and systems (iros)},
  pages={7827--7834},
  year={2021},
  organization={IEEE}
}

@article{bjorck2025gr00t,
  title={Gr00t n1: An open foundation model for generalist humanoid robots},
  author={Bjorck, Johan and Casta{\~n}eda, Fernando and Cherniadev, Nikita and Da, Xingye and Ding, Runyu and Fan, Linxi and Fang, Yu and Fox, Dieter and Hu, Fengyuan and Huang, Spencer and others},
  journal={arXiv preprint arXiv:2503.14734},
  year={2025}
}

@article{mccarthy2025towards,
  title={Towards generalist robot learning from internet video: A survey},
  author={McCarthy, Robert and Tan, Daniel CH and Schmidt, Dominik and Acero, Fernando and Herr, Nathan and Du, Yilun and Thuruthel, Thomas G and Li, Zhibin},
  journal={Journal of Artificial Intelligence Research},
  volume={83},
  year={2025}
}

@article{labbe2022megapose,
  title={Megapose: 6d pose estimation of novel objects via render \& compare},
  author={Labb{\'e}, Yann and Manuelli, Lucas and Mousavian, Arsalan and Tyree, Stephen and Birchfield, Stan and Tremblay, Jonathan and Carpentier, Justin and Aubry, Mathieu and Fox, Dieter and Sivic, Josef},
  journal={arXiv preprint arXiv:2212.06870},
  year={2022}
}

@article{du2023learning,
  title={Learning universal policies via text-guided video generation},
  author={Du, Yilun and Yang, Sherry and Dai, Bo and Dai, Hanjun and Nachum, Ofir and Tenenbaum, Josh and Schuurmans, Dale and Abbeel, Pieter},
  journal={Advances in neural information processing systems},
  volume={36},
  pages={9156--9172},
  year={2023}
}

@article{tian2024predictive,
  title={Predictive inverse dynamics models are scalable learners for robotic manipulation},
  author={Tian, Yang and Yang, Sizhe and Zeng, Jia and Wang, Ping and Lin, Dahua and Dong, Hao and Pang, Jiangmiao},
  journal={arXiv preprint arXiv:2412.15109},
  year={2024}
}

@article{team2025evaluating,
  title={Evaluating Gemini Robotics Policies in a Veo World Simulator},
  author={Team, Gemini Robotics and Choromanski, Krzysztof and Devin, Coline and Du, Yilun and Dwibedi, Debidatta and Gao, Ruiqi and Jindal, Abhishek and Kipf, Thomas and Kirmani, Sean and Leal, Isabel and others},
  journal={arXiv preprint arXiv:2512.10675},
  year={2025}
}

@article{luo2024grounding,
  title={Grounding video models to actions through goal conditioned exploration},
  author={Luo, Yunhao and Du, Yilun},
  journal={arXiv preprint arXiv:2411.07223},
  year={2024}
}

@article{yang2023learning,
  title={Learning interactive real-world simulators},
  author={Yang, Mengjiao and Du, Yilun and Ghasemipour, Kamyar and Tompson, Jonathan and Schuurmans, Dale and Abbeel, Pieter},
  journal={arXiv preprint arXiv:2310.06114},
  volume={1},
  number={2},
  pages={6},
  year={2023}
}

@article{wu2025hunyuanvideo,
  title={Hunyuanvideo 1.5 technical report},
  author={Wu, Bing and Zou, Chang and Li, Changlin and Huang, Duojun and Yang, Fang and Tan, Hao and Peng, Jack and Wu, Jianbing and Xiong, Jiangfeng and Jiang, Jie and others},
  journal={arXiv preprint arXiv:2511.18870},
  year={2025}
}

@article{team2025klingavatar,
  title={Klingavatar 2.0 technical report},
  author={Team, Kling and Chen, Jialu and Ding, Yikang and Fang, Zhixue and Gai, Kun and Gao, Yuan and He, Kang and Hua, Jingyun and Jiang, Boyuan and Lao, Mingming and others},
  journal={arXiv preprint arXiv:2512.13313},
  year={2025}
}

@article{hong2022cogvideo,
  title={Cogvideo: Large-scale pretraining for text-to-video generation via transformers},
  author={Hong, Wenyi and Ding, Ming and Zheng, Wendi and Liu, Xinghan and Tang, Jie},
  journal={arXiv preprint arXiv:2205.15868},
  year={2022}
}

@article{wang2025lavie,
  title={Lavie: High-quality video generation with cascaded latent diffusion models},
  author={Wang, Yaohui and Chen, Xinyuan and Ma, Xin and Zhou, Shangchen and Huang, Ziqi and Wang, Yi and Yang, Ceyuan and He, Yinan and Yu, Jiashuo and Yang, Peiqing and others},
  journal={International Journal of Computer Vision},
  volume={133},
  number={5},
  pages={3059--3078},
  year={2025},
  publisher={Springer}
}

@article{hacohen2024ltx,
  title={Ltx-video: Realtime video latent diffusion},
  author={HaCohen, Yoav and Chiprut, Nisan and Brazowski, Benny and Shalem, Daniel and Moshe, Dudu and Richardson, Eitan and Levin, Eran and Shiran, Guy and Zabari, Nir and Gordon, Ori and others},
  journal={arXiv preprint arXiv:2501.00103},
  year={2024}
}

@misc{isaacsim2023,
  title        = {NVIDIA Isaac Sim: High-Fidelity Simulation for Robotics},
  author       = {{NVIDIA Corporation}},
  year         = {2023},
  howpublished = {\url{https://developer.nvidia.com/isaac-sim}},
  note         = {Accessed: 2026-03-03}
}

@article{agarwal2025cosmos,
  title={Cosmos world foundation model platform for physical ai},
  author={Agarwal, Niket and Ali, Arslan and Bala, Maciej and Balaji, Yogesh and Barker, Erik and Cai, Tiffany and Chattopadhyay, Prithvijit and Chen, Yongxin and Cui, Yin and Ding, Yifan and others},
  journal={arXiv preprint arXiv:2501.03575},
  year={2025}
}

@inproceedings{wu2025foundation,
  title={Foundation Feature-Driven Online End-Effector Pose Estimation: A Marker-Free and Learning-Free Approach},
  author={Wu, Tianshu and Zhang, Jiyao and Liang, Shiqian and Han, Zhengxiao and Dong, Hao},
  booktitle={2025 IEEE International Conference on Robotics and Automation (ICRA)},
  pages={1921--1928},
  year={2025},
  organization={IEEE}
}

@article{shang2026worldarena,
  title={WorldArena: A Unified Benchmark for Evaluating Perception and Functional Utility of Embodied World Models},
  author={Shang, Yu and Li, Zhuohang and Ma, Yiding and Su, Weikang and Jin, Xin and Wang, Ziyou and Jin, Lei and Zhang, Xin and Tang, Yinzhou and Su, Haisheng and others},
  journal={arXiv preprint arXiv:2602.08971},
  year={2026}
}

@inproceedings{yang2024depth,
  title={Depth anything: Unleashing the power of large-scale unlabeled data},
  author={Yang, Lihe and Kang, Bingyi and Huang, Zilong and Xu, Xiaogang and Feng, Jiashi and Zhao, Hengshuang},
  booktitle={Proceedings of the IEEE/CVF conference on computer vision and pattern recognition},
  pages={10371--10381},
  year={2024}
}

@inproceedings{li2025megasam,
  title={Megasam: Accurate, fast and robust structure and motion from casual dynamic videos},
  author={Li, Zhengqi and Tucker, Richard and Cole, Forrester and Wang, Qianqian and Jin, Linyi and Ye, Vickie and Kanazawa, Angjoo and Holynski, Aleksander and Snavely, Noah},
  booktitle={Proceedings of the IEEE/CVF Conference on Computer Vision and Pattern Recognition},
  pages={10486--10496},
  year={2025}
}

@inproceedings{chen2025video,
  title={Video depth anything: Consistent depth estimation for super-long videos},
  author={Chen, Sili and Guo, Hengkai and Zhu, Shengnan and Zhang, Feihu and Huang, Zilong and Feng, Jiashi and Kang, Bingyi},
  booktitle={Proceedings of the Computer Vision and Pattern Recognition Conference},
  pages={22831--22840},
  year={2025}
}

@article{Qwen3-VL,
      title={Qwen3-VL Technical Report}, 
      author={Shuai Bai and Yuxuan Cai and Ruizhe Chen and Keqin Chen and Xionghui Chen and Zesen Cheng and Lianghao Deng and Wei Ding and Chang Gao and Chunjiang Ge and Wenbin Ge and Zhifang Guo and Qidong Huang and Jie Huang and Fei Huang and Binyuan Hui and Shutong Jiang and Zhaohai Li and Mingsheng Li and Mei Li and Kaixin Li and Zicheng Lin and Junyang Lin and Xuejing Liu and Jiawei Liu and Chenglong Liu and Yang Liu and Dayiheng Liu and Shixuan Liu and Dunjie Lu and Ruilin Luo and Chenxu Lv and Rui Men and Lingchen Meng and Xuancheng Ren and Xingzhang Ren and Sibo Song and Yuchong Sun and Jun Tang and Jianhong Tu and Jianqiang Wan and Peng Wang and Pengfei Wang and Qiuyue Wang and Yuxuan Wang and Tianbao Xie and Yiheng Xu and Haiyang Xu and Jin Xu and Zhibo Yang and Mingkun Yang and Jianxin Yang and An Yang and Bowen Yu and Fei Zhang and Hang Zhang and Xi Zhang and Bo Zheng and Humen Zhong and Jingren Zhou and Fan Zhou and Jing Zhou and Yuanzhi Zhu and Ke Zhu},
	  journal={arXiv preprint arXiv:2511.21631},
      year={2025}
}

\newpage
\appendix

\section{Limitations}

RoboWM-Bench evaluates predicted manipulation videos by converting them into executable actions and validating them through simulation-based execution. Although we validate both the real-to-sim environments and the action-extraction modules, as discussed in Section~\ref{subsec:validation}, the protocol still relies on intermediate video-to-action interfaces. Such errors are most likely to affect highly contact-sensitive or near-boundary interactions, where small pose or grasp deviations can change execution outcomes. For human-hand videos, RoboWM-Bench assesses executability through retargeted robot actions rather than direct simulation of human-hand dynamics, which remain challenging to model faithfully in current physics engines. While the current task suite covers diverse tabletop manipulation settings, it is not exhaustive; future extensions could include broader scene layouts and object categories. Finally, the step-level evaluation relies on task-specific key nodes and manually specified checkers. These checkers provide interpretable diagnostics, but scaling to larger task families may benefit from LLM- or VLM-assisted generation of simulation-based success predicates and key-node checkers, while retaining execution as the final validation signal.
\section{Additional Results for Purely Simulated Robotic Tasks}
\label{sec:robot_sim}

RoboWM-Bench also includes a set of robotic manipulation tasks evaluated entirely in simulation environments. 

The task setup follows the same protocol described in the main paper. For each task, video world models generate future manipulation behaviors conditioned on the simulation observations and the corresponding task descriptions. 
The predicted behaviors are then converted into executable action sequences and executed in simulation to assess task completion.

Table~\ref{tab:supp_robot_sim} reports both task-level and step-level execution success rates for these purely simulated tasks. 
The overall trends are consistent with those observed in the real-to-sim evaluation. As task complexity increases, the success rates of most models decrease, particularly for tasks requiring long-horizon reasoning or precise contact interactions. 


\begin{table*}[htbp]
\caption{Task-level and step-level embodied execution success rates (\%) on RoboWM-Bench for robotic manipulation tasks evaluated entirely in simulation.}	
\label{tab:supp_robot_sim}	
\centering
{\scriptsize
\setlength{\tabcolsep}{3pt}
\renewcommand{\arraystretch}{0.8}
\begin{tabularx}{\textwidth}{p{1.8cm}|*{6}{|>{\centering\arraybackslash}X}}
\toprule
\textbf{Method} & \multicolumn{6}{c}{\textbf{Robot (Task Level)}} \\
\cmidrule(lr){2-7}
& \scriptsize Close Drawer & \scriptsize Push Button & \scriptsize Cut Sausage & \scriptsize Turn Off Faucet & \scriptsize Assemble Burger & \scriptsize Fold Clothes \\
\midrule    
Cosmos    & 0\%  & 10\% & 10\% & 0\% & 0\% & 0\%  \\
Wan 2.2   & 10\% & 10\% & 10\%  & 0\%  & 0\%  & 0\%  \\
Wan 2.6   & 30\% & 20\% & 40\% & 0\% & 0\% & 0\% \\
Veo       & 10\% & 20\% & 20\% & 0\% & 0\% & 0\%  \\
\bottomrule
\end{tabularx}

\vspace{2mm} 

\begin{tabularx}{\textwidth}{p{1.8cm}|*{8}{|>{\centering\arraybackslash}X}}
\toprule
& \multicolumn{8}{c}{\textbf{Robot (Step Level)}} \\
\cmidrule(lr){2-9}
\textbf{Method} & \multicolumn{3}{c|}{\textbf{Turn Off Faucet}} & \multicolumn{3}{c|}{\textbf{Assemble Burger}}  & \multicolumn{2}{c}{\textbf{Fold Clothes}} \\
& contact & rot. $> 20^{\circ}$ & turn off & contact & lift & place & L. sleeve & R. sleeve \\
\midrule
Cosmos    & 10\% & 0\% & 0\% & 10\% & 0\% & 0\% & 0\% & 0\% \\
Wan 2.2   & 10\% & 0\% & 0\% & 10\% & 0\% & 0\% & 0\% & 0\% \\
Wan 2.6   & 40\% & 10\% & 0\% & 30\% & 0\% & 0\% & 0\% & 0\% \\
Veo       & 20\% & 0\% & 0\% & 10\% & 0\% & 0\% & 0\% & 0\% \\
\bottomrule
\end{tabularx}

}
\end{table*}
\section{Additional Results on Human Tasks}
\label{sec:bimanual}

We additionally include human-hand manipulation tasks involving bimanual coordination.
The detailed execution success rates are reported in Table~\ref{tab:supp_human}.
The results show that bimanual tasks are generally more challenging, as successful execution requires coordinated interactions between both hands to maintain physically feasible manipulation.

\begin{table*}[htbp]
\caption{
Task-level and step-level embodied execution success rates (\%) on the additional human manipulation tasks involving bimanual coordination.
}	
\label{tab:supp_human}	
\centering
{\scriptsize
\setlength{\tabcolsep}{3pt}
\renewcommand{\arraystretch}{0.8}

\begin{tabularx}{\textwidth}{l|c|>{\centering\arraybackslash}X|>{\centering\arraybackslash}X|>{\centering\arraybackslash}X|>{\centering\arraybackslash}X|>{\centering\arraybackslash}X|>{\centering\arraybackslash}X|>{\centering\arraybackslash}X}
\toprule

& \multicolumn{1}{c|}{\textbf{Task Level}} & \multicolumn{7}{c}{\textbf{Step Level}} \\
\cmidrule(lr){2-2} \cmidrule(lr){3-9}

\textbf{Method} & \multicolumn{1}{c|}{\textbf{Cook}}
& grasp spatula
& grasp pan
& lift spatula
& lift pan
& spatula--pan contact
& place spatula
& place pan \\
\midrule

Cosmos  & 10\% & 30\% & 20\% & 20\% & 20\% & 20\% & 10\% & 20\% \\
Wan 2.2 & 30\% & 70\% & 60\% & 40\% & 30\% & 30\% & 30\% & 30\% \\
Wan 2.6 & \textbf{50\%} & \textbf{90\%} & \textbf{70\%} & \textbf{70\%} & \textbf{70\%} & \textbf{60\%} & \textbf{50\%} & \textbf{60\%} \\
Veo & 40\% & 80\% & \textbf{70\%} & \textbf{70\%} & 50\% & 60\% & 40\% & 50\% \\
LVP     & 30\% & 80\% & \textbf{70\%} & 60\% & 50\% & 40\% & 40\% & 30\% \\

\midrule
\midrule

& \multicolumn{1}{c|}{\textbf{Task Level}} & \multicolumn{4}{c|}{\textbf{Step Level}} \\
\cmidrule(lr){2-2} \cmidrule(lr){3-6}
\textbf{Method} & \multicolumn{1}{c|}{\textbf{Lift Large Box}}
& grasp left
& grasp right
& lift
& place
& \multicolumn{1}{c}{}
& \multicolumn{1}{c}{}
& \\
\midrule

Cosmos  & 10\% & 20\% & 20\% & 10\% & 10\% & \multicolumn{1}{c}{} & \multicolumn{1}{c}{} & \multicolumn{1}{c}{} \\
Wan 2.2 & 30\% & 70\% & 60\% & 30\% & 30\% & \multicolumn{1}{c}{} & \multicolumn{1}{c}{} & \multicolumn{1}{c}{} \\
Wan 2.6 & \textbf{60\%} & \textbf{90\%} & \textbf{90\%} & \textbf{70\%} & \textbf{60\%} & \multicolumn{1}{c}{} & \multicolumn{1}{c}{} & \multicolumn{1}{c}{}  \\
Veo & 50\% & 80\% & 80\% & 50\% & 50\% & \multicolumn{1}{c}{} & \multicolumn{1}{c}{} & \multicolumn{1}{c}{}  \\
LVP     & 30\% & 70\% & 70\% & 30\% & 30\% & \multicolumn{1}{c}{} & \multicolumn{1}{c}{} & \multicolumn{1}{c}{}  \\

\bottomrule
\end{tabularx}






}
\end{table*}
\section{Comparison with PAI-Bench Quality Scores}
\label{sec:pai_quality_score}

\begin{figure*}[htbp]
\centering
\includegraphics[width=0.9\linewidth]{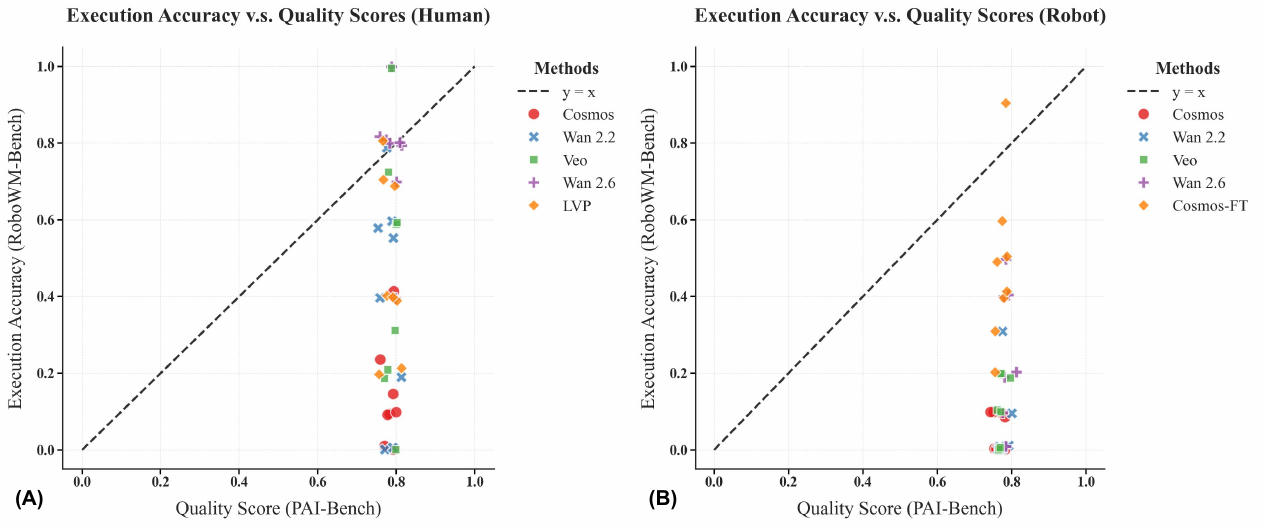}
\caption{
Comparison between the average quality scores in PAI-Bench with the execution accuracy in RoboWM-Bench. The left scatter plot shows human-hand tasks, and the right scatter plot shows robotic tasks. 
} 
\label{fig:supp_pai_quality_score}
\end{figure*}

We additionally compare the average quality scores in PAI-Bench~\cite{zhou2025pai} with the execution accuracy in RoboWM-Bench, as shown in Figure~\ref{fig:supp_pai_quality_score}. Specifically, Figure~\ref{fig:supp_pai_quality_score} presents two scatter plots corresponding to human-hand tasks and robotic tasks, respectively. In both plots, the horizontal axis denotes the average quality score in PAI-Bench, while the vertical axis represents the execution accuracy measured in RoboWM-Bench. Each point corresponds to a specific world model baseline (indicated by color) on a particular manipulation task, reporting its average quality score and execution accuracy. 

As observed in the scatter plots, most points cluster along a vertical line around an average quality score of approximately 0.78, suggesting that the PAI-Bench quality scores are relatively consistent across different tasks and models.
In contrast, the execution accuracy measured by RoboWM-Bench exhibits substantially greater variation. This discrepancy suggests that visual plausibility does not necessarily imply physical correctness, and the embodiment-grounded evaluation in RoboWM-Bench provides a more sensitive and informative measure of physical executability.

Following the definitions and reporting protocol of PAI-Bench, we additionally present detailed domain and quality scores in tabular form. 
Note that all scores are computed using videos generated for the manipulation tasks in RoboWM-Bench, rather than the broader video sources used in PAI-Bench.
Specifically, the detailed domain scores are reported in Table~\ref{tab:supp_domain_human} for human-hand tasks and Table~\ref{tab:supp_domain_robot} for robotic tasks, while the detailed quality scores are reported in Table~\ref{tab:supp_quality_human} and Table~\ref{tab:supp_quality_robot}, respectively.


\begin{table*}[htbp]
    \centering
    \caption{\textbf{PAI-Bench Domain Score} on Human-Hand Tasks.}
    \label{tab:supp_domain_human}
    
    \small 
    \setlength{\tabcolsep}{3pt} 
    \renewcommand{\arraystretch}{1} 

    \newcolumntype{Y}{>{\centering\arraybackslash}X} 
    \newcolumntype{L}{>{\raggedright\arraybackslash}p{1.7cm}} 
    \newcolumntype{A}{>{\centering\arraybackslash}p{0.7cm}} 

    \begin{tabularx}{\textwidth}{L | *{8}{Y|} A}  
        \toprule
        \textbf{Models} & \multicolumn{8}{c|}{\textbf{Domain Score}} & \textbf{Avg.} \\  
        \cline{2-9}  
        & \scriptsize Pick Object & \scriptsize Push Button & \scriptsize Put on Plate & \scriptsize Pour Water & \scriptsize Stack Cups &  \scriptsize Open Drawer & \scriptsize Put in Drawer & \scriptsize Fold Towel
        \\  
        \midrule
        Cosmos    & 96 & \textbf{100} & 92 & 72 & 78 & \textbf{100} & \textbf{100} & 86 & 90.5 \\
        Wan 2.2   & 92 & \textbf{100} & \textbf{100} & 94 & \textbf{100} & \textbf{100} & \textbf{100} & \textbf{100} & \textbf{98.3} \\
        Wan 2.6   & \textbf{100} & 92 & \textbf{100} & \textbf{100} & \textbf{100} & \textbf{100} & \textbf{100} & 86 & 97.3 \\
        Veo       & \textbf{100} & 82 & \textbf{100} & \textbf{100} & \textbf{100} & 92 & 92 & \textbf{100} & 95.8 \\
        LVP       & \textbf{100} & 78 & \textbf{100} & 78 & \textbf{100} & \textbf{100} & 86 & 76 & 87.0 \\
        \bottomrule
    \end{tabularx}
\end{table*}

\begin{table*}[htbp]
    \centering
\caption{\textbf{PAI-Bench Domain Score} on Robotic Real Tasks.}
\label{tab:supp_domain_robot}

\small 
\setlength{\tabcolsep}{3pt} 
\renewcommand{\arraystretch}{1.0} 

\newcolumntype{Y}{>{\centering\arraybackslash}X} 
\newcolumntype{L}{>{\raggedright\arraybackslash}p{1.7cm}} 
\newcolumntype{A}{>{\centering\arraybackslash}p{0.7cm}} 

\begin{tabularx}{\textwidth}{L | *{8}{Y|} A}  
    \toprule
    \textbf{Models} & \multicolumn{8}{c|}{\textbf{Domain Scores}} & \textbf{Avg.} \\  
    \cline{2-9}  
    & \scriptsize Close Drawer & \scriptsize Pick Object & \scriptsize Push Object & \scriptsize Push Button & \scriptsize Put on Plate & \scriptsize Discard Trash & \scriptsize Pull Object & \scriptsize Put in Drawer & \\  
    \midrule
    Cosmos    & 84 & \textbf{100} & \textbf{100} & \textbf{100} & \textbf{100} & \textbf{100} & \textbf{100} & \textbf{100} & 98.0 \\
    Wan 2.2   & \textbf{100} & \textbf{100} & 92 & \textbf{100} & 94 & \textbf{100} & 98 & 86 & 96.3 \\
    Wan 2.6   & \textbf{100} & \textbf{100} & \textbf{100} & \textbf{100} & \textbf{100} & \textbf{100} & \textbf{100} & \textbf{100} & \textbf{99.5} \\
    Veo       & \textbf{100} & \textbf{100} & \textbf{100} & 90 & \textbf{100} & \textbf{100} & \textbf{100} & 80 & 96.3 \\
    Cosmos-FT & \textbf{100} & \textbf{100} & \textbf{100} & \textbf{100} & \textbf{100} & 92 & \textbf{100} & 100 & \textbf{99.0} \\
    \bottomrule
\end{tabularx}
\end{table*}

\begin{table*}[htbp]
    \centering
\caption{\textbf{PAI-Bench Quality Score} on Human-Hand Tasks. Metrics follow the PAI-Bench protocol, including Subject Consistency (SC), Background Consistency (BC), Motion Smoothness (MS), Aesthetic Quality (AQ), Imaging Quality (IQ), Overall Consistency (OC), I2V Subject (IS), I2V Background (IB).}
\label{tab:supp_quality_human}

\small 
\setlength{\tabcolsep}{3pt} 
\renewcommand{\arraystretch}{1.0} 

\newcolumntype{Y}{>{\centering\arraybackslash}X} 
\newcolumntype{L}{>{\raggedright\arraybackslash}p{1.7cm}} 
\newcolumntype{A}{>{\centering\arraybackslash}p{0.7cm}} 

\begin{tabularx}{0.8\textwidth}{L | *{8}{Y|} A}  
    \toprule
    \textbf{Models} & \multicolumn{8}{c|}{\textbf{Quality Score}} & \textbf{Avg.} \\  
    \cline{2-9}  
    & \textbf{SC} & \textbf{BC} & \textbf{MS} & \textbf{AQ} & \textbf{IQ} & \textbf{OC} & \textbf{IS} & \textbf{IB} & \\  
    \midrule
    Cosmos    & 95.0 & 94.9 & 99.5 & \textbf{45.8} & 74.9 & 25.6 & 98.3 & 98.5 & 79.1 \\
    Wan 2.2   & 95.5 & 94.9 & 99.1 & 41.7 & 74.5 & 26.2 & \textbf{98.6} & 98.5 & 78.6 \\
    Wan 2.6   & \textbf{96.9} & 95.6 & 99.2 & 40.5 & \textbf{75.9} & 25.8 & 98.4 & 97.9 & 78.8 \\
    Veo       & 96.0 & 95.5 & \textbf{99.6} & 40.9 & 75.4 & \textbf{26.3} & 98.5 & \textbf{98.6} & \textbf{78.9} \\
    LVP       & 96.3 & \textbf{95.9} & 99.4 & 36.4 & 74.4 & 25.6 & 98.1 & 98.0 & 78.0 \\
    
    \bottomrule
\end{tabularx}
\end{table*}

\begin{table*}[htbp]
    \centering
\caption{\textbf{PAI-Bench Quality Score} on Robotic Real Tasks. }
\label{tab:supp_quality_robot}

\small 
\setlength{\tabcolsep}{3pt} 
\renewcommand{\arraystretch}{1.0} 

\newcolumntype{Y}{>{\centering\arraybackslash}X} 
\newcolumntype{L}{>{\raggedright\arraybackslash}p{1.7cm}} 
\newcolumntype{A}{>{\centering\arraybackslash}p{0.7cm}} 

\begin{tabularx}{0.8\textwidth}{L | *{8}{Y|} A}  
    \toprule
    \textbf{Models} & \multicolumn{8}{c|}{\textbf{Quality Score}} & \textbf{Avg.} \\  
    \cline{2-9}  
    & \textbf{SC} & \textbf{BC} & \textbf{MS} & \textbf{AQ} & \textbf{IQ} & \textbf{OC} & \textbf{IS} & \textbf{IB} & \\  
    \midrule
    Cosmos    & 94.9 & 92.9 & 99.4 & 44.3 & 74.5 & 22.2 & 94.0 & 93.5 & 77.0 \\
    Wan 2.2   & 96.5 & 94.6 & 99.5 & 43.9 & 73.5 & 23.2 & 94.6 & 93.2 & 77.4 \\
    Wan 2.6   & \textbf{96.9} & \textbf{96.2} & 99.4 & \textbf{45.1} & \textbf{77.1} & 23.3 & 94.3 & 92.8 & \textbf{78.1} \\
    Veo       & 95.1 & 92.5 & \textbf{99.6} & 44.4 & 70.1 & \textbf{23.6} & 95.0 & 94.8 & 76.9 \\
    Cosmos-FT & 94.6 & 94.2 & 99.5 & 44.3 & 64.9 & 22.8 & \textbf{98.2} & \textbf{98.5} & 77.1 \\
    \bottomrule
\end{tabularx}
\end{table*}

\newpage
\section{More Visualizations}
\label{sec:more_visu}

\begin{figure*}[htbp]
\centering
\includegraphics[width=\linewidth]{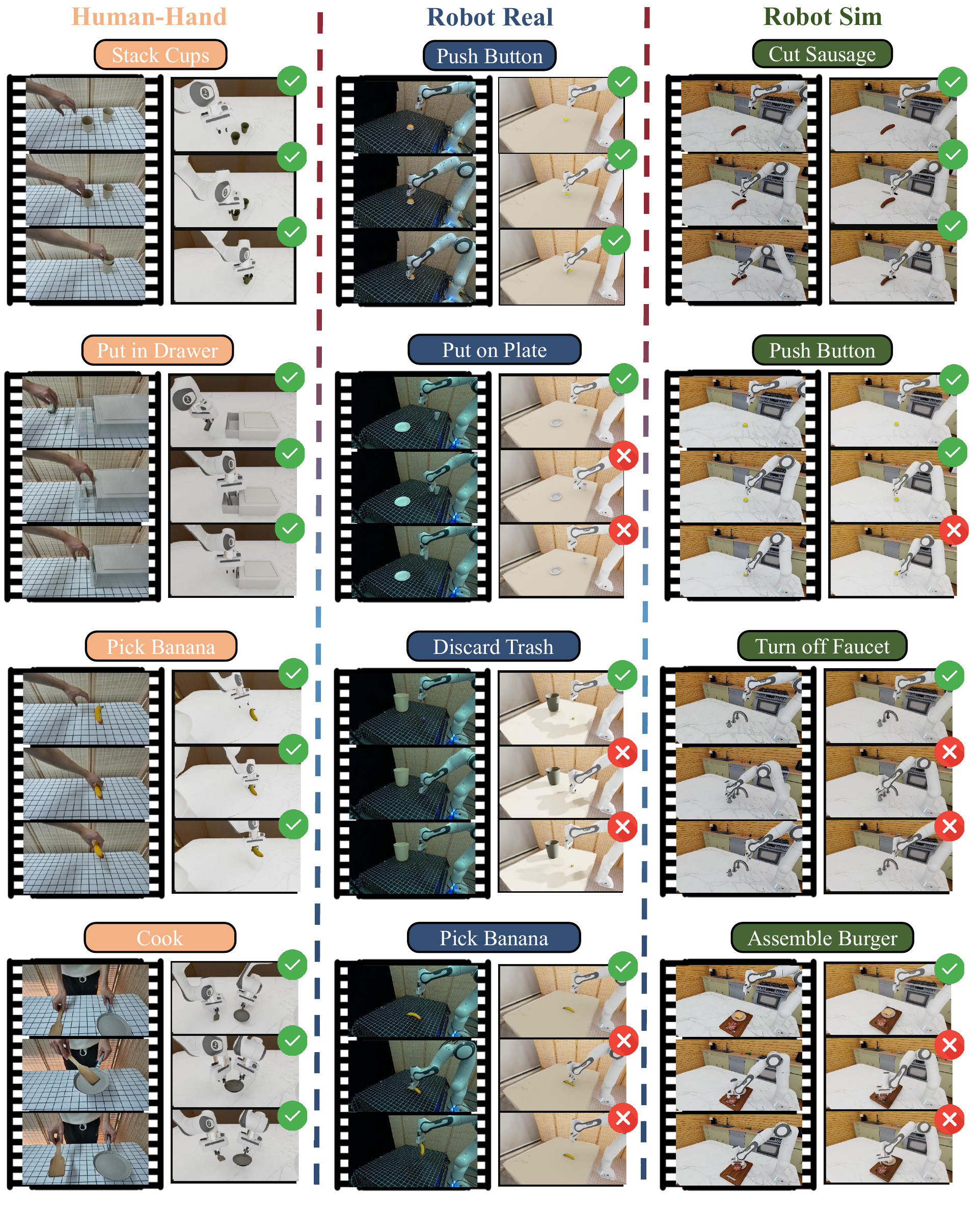}
\caption{
\textbf{Additional qualitative results on RoboWM-Bench.} For each task, predicted videos (left) are converted into robot actions and executed in simulation (right).
} 
\label{fig:supp_results}
\end{figure*}

Figure~\ref{fig:supp_results} provides additional qualitative examples of the predicted manipulation videos and their corresponding embodied executions. These results complement the examples shown in Figure~3 of the main paper.


\section{Discussion of Depth in Human-Hand Tracking}
\label{sec:depth}


In our experiments, we empirically found that Phantom~\cite{lepert2503phantom} provides the most reliable performance for pose tracking and retargeting in human-hand manipulation videos. Therefore, as described in Section~3.3.1 of the main paper, our pipeline builds upon Phantom with several adaptations.

However, Phantom assumes access to ground-truth depth information, whereas in our setting such depth is not available for the videos predicted by world models. In practice, only the depth of the first frame can be obtained, as it is captured by the camera in the real-world or simulation environment.

\begin{figure*}[htbp]
\centering
\includegraphics[width=\linewidth]{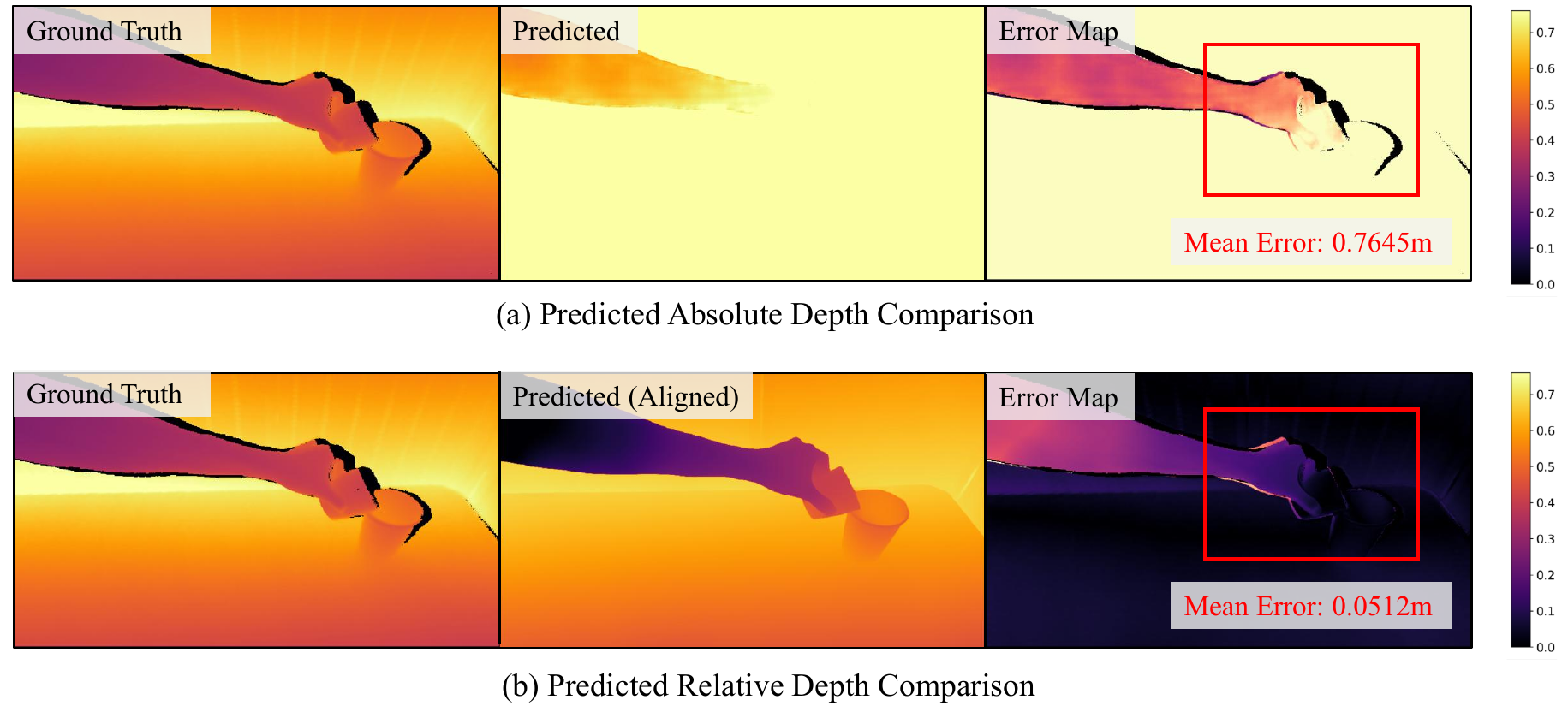}
\caption{
Visualization of depth estimation results.
(a) The predicted absolute depth shows a large discrepancy from the ground-truth.
(b) Aligning relative depth with the first-frame ground-truth depth improves consistency, although non-negligible errors still remain.
}
\label{fig:supp_depth}
\end{figure*}

Recent works have explored leveraging depth estimation for video understanding~\cite{chen2025large,li2025megasam}. In our setting, we investigate whether estimated depth can improve human-hand tracking.
To this end, we estimate depth from RGB frames using Video Depth Anything~\cite{yang2024depth,chen2025video} and evaluate two strategies. 
First, we directly use the absolute depth predicted by the model. However, as shown in Figure~\ref{fig:supp_depth}(a), the predicted depth shows a large discrepancy from the ground-truth. 
Second, we estimate relative depth for subsequent frames and align it with the ground-truth depth of the first frame to recover absolute depth values. As illustrated in Figure~\ref{fig:supp_depth}(b), the recovered depth remains imperfect.

Empirically, we observe that incorporating these estimated depths does not improve downstream pose tracking and retargeting performance. Therefore, depth information is not used in the final pipeline.

\section{Compute Resources}
\label{app:compute}

The runtime of video-to-action processing depends on the video resolution, frame rate, and duration. We report representative processing times and GPU memory usage for the two procedures used in our evaluation. For human-hand pose estimation and retargeting, processing a predicted video at 1280$\times$720 resolution and 15 FPS, with a duration of up to 25 seconds, took approximately 112 seconds on a single NVIDIA RTX 4090 GPU and used about 5.5 GB of GPU memory. For IDM inference on robotic manipulation videos, processing a 1280$\times$720, 30-FPS, 5-second video took approximately 1.5 seconds on a single NVIDIA A800 GPU and used about 4 GB of GPU memory. The released code includes scripts for reproducing the evaluation on user hardware.

\section{Broader impacts}
RoboWM-Bench aims to support safer and more reliable development of video world models for robotic manipulation by providing standardized, reproducible, and execution-grounded evaluation. Its positive impact lies in revealing cases where visually plausible generated videos may not correspond to executable manipulation behaviors. Potential risks include overinterpreting benchmark scores as sufficient evidence for real-world robot deployment or applying video-to-action pipelines without appropriate safeguards. To mitigate these risks, RoboWM-Bench is intended as a diagnostic evaluation protocol, with validation conducted in simulation or real-to-sim environments and without releasing personally identifiable or sensitive data.

\section{Prompt Details for World Model Video Generation}
In this section, we provide details on how instruction prompts are generated for world model video generation.

\subsection{Human-Hand Tasks}
To generate instruction prompts for video world models on human tasks, we leverage the Qwen3-VL-Flash \cite{Qwen3-VL} model to generate concise task-specific descriptions conditioned on the initial observation image and high-level task instructions (e.g., \emph{Pick up the stapler on the table, and stay still.}). The resulting descriptions generated by Qwen are summarized in Table~\ref{tab:human_prompts} and Table~\ref{tab:human_bi_prompts}.

To further encourage physical consistency of the synthesized videos, we append a standardized set of constraints to the generated instruction: \textit{"The entire human hand must remain fully visible in the frame at all times, with no cropping, no fingers cut off, and no part of the hand outside the camera view. The camera must remain completely static with no movement, no panning, no tilting, no zooming, and no change in viewpoint during the entire video. No extra or unnecessary motions are allowed, the hand must not perform any additional gestures such as turning to show the palm, posing, rotating unnecessarily. All other objects in the scene that are not being manipulated must remain completely stationary, with no position shift, no rotation, and no change in placement throughout the entire video."} Empirically, we observe that including these constraints improves the quality of videos generated by the world models.



\begin{table*}[htbp]
    
    \centering
    \small 
    
    \setlength{\aboverulesep}{0pt}
    \setlength{\belowrulesep}{0pt}
    \setlength{\tabcolsep}{8pt}    
    \renewcommand{\arraystretch}{1.4} 

    \caption{Overview of human-hand manipulation tasks and their corresponding instruction prompts.}
    \label{tab:human_prompts}	
    \vspace{0.8em}

    \newcolumntype{K}{>{\centering\arraybackslash}p{2.2cm}}
    
    \newcolumntype{M}{>{\centering\arraybackslash}p{0.28\textwidth}}
    
    \renewcommand{\tabularxcolumn}[1]{p{#1}} 
    \newcolumntype{L}{>{\raggedright\arraybackslash\scriptsize}X} 

    \begin{tabularx}{\textwidth}{K | M | L}
    \hline
    \textbf{Task Name} & \textbf{Input Image} & \multicolumn{1}{>{\centering\arraybackslash}X}{\textbf{Task Prompt}} \\
    \hline

    Pick Object & 
    \includegraphics[width=0.24\textwidth, valign=t, margin=0pt 6pt 0pt 6pt]{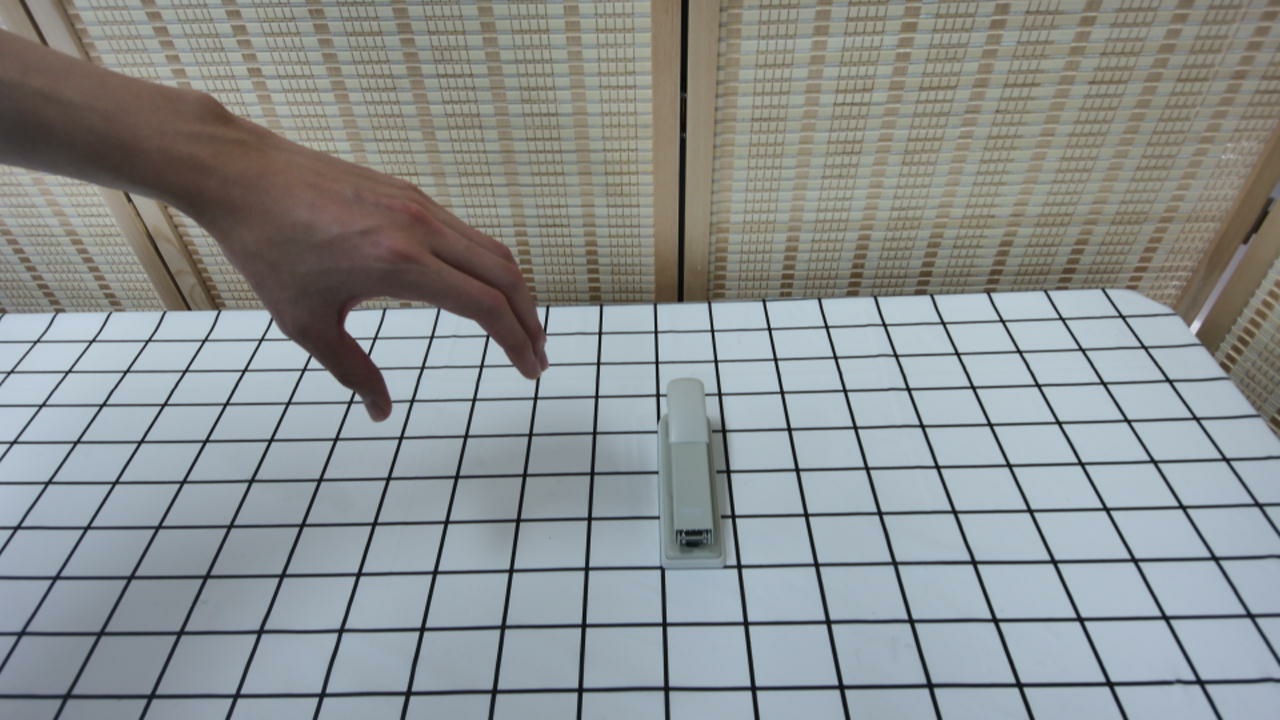} & 
    The hand lowers slowly toward the stapler and gently grasps it using the thumb and index finger. The stapler is lifted slowly from the table surface without any bouncing or sliding. \\
    
    \hline

    Push Button & 
    \includegraphics[width=0.24\textwidth, valign=t, margin=0pt 6pt 0pt 6pt]{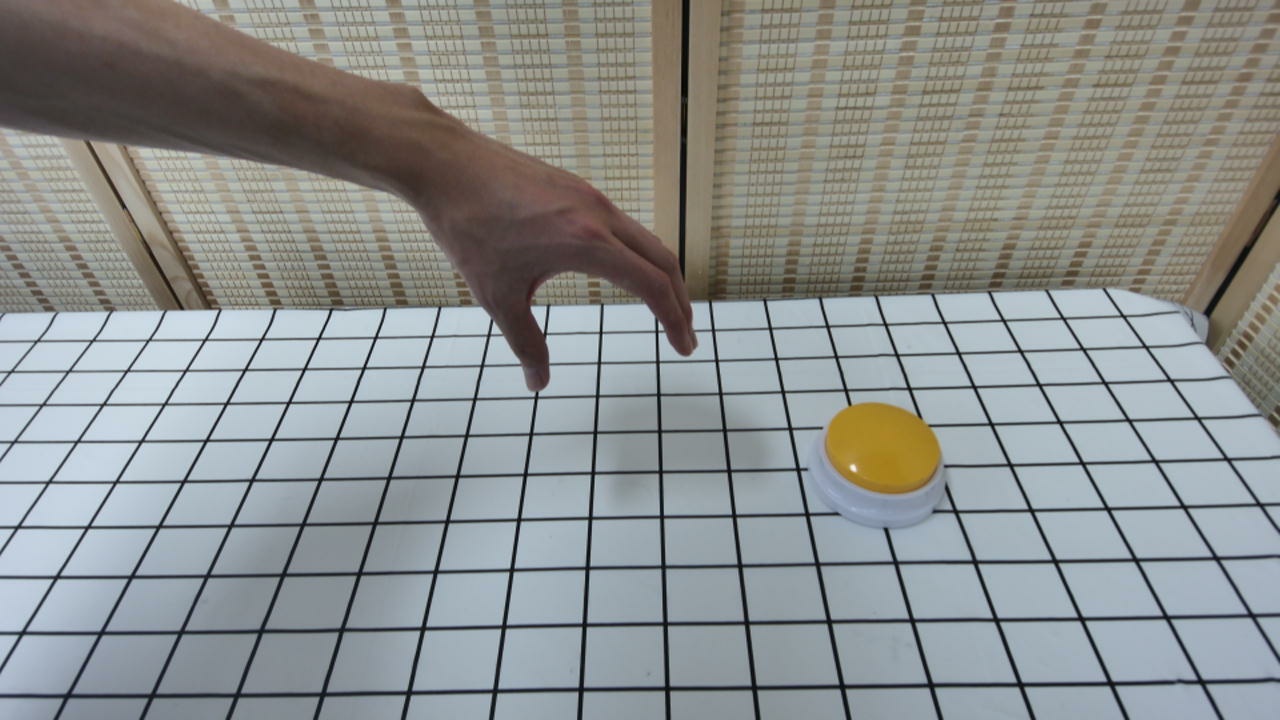} & 
    The hand moves downward toward the button. The fingers make contact with the yellow surface. The hand applies pressure causing the button to compress. The hand lifts up and rest in the air. \\
    
    \hline

    Put on Plate & 
    \includegraphics[width=0.24\textwidth, valign=t, margin=0pt 6pt 0pt 6pt]{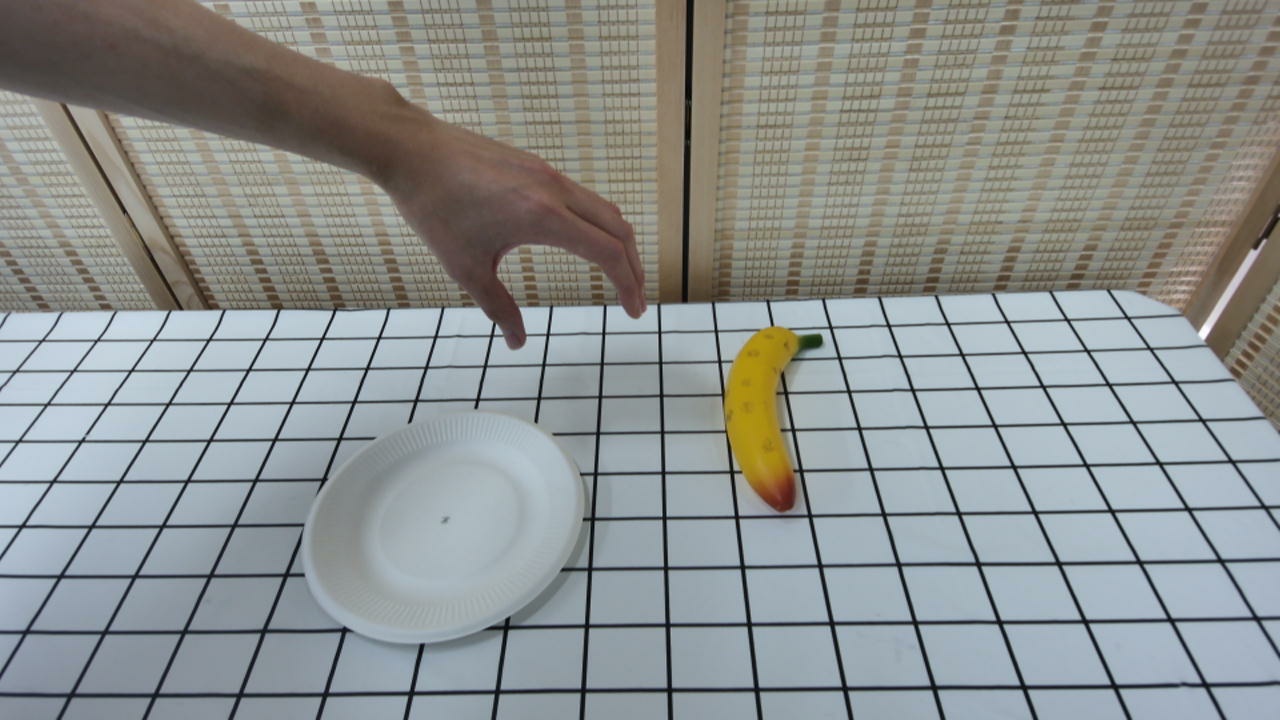} & 
    The hand slowly extend downward to gently grasp the banana using the thumb and index finger and then slowly lifts the banana straight up. The banana is then slowly placed on the center of the plate. \\

    \hline

    Pour Water & 
    \includegraphics[width=0.24\textwidth, valign=t, margin=0pt 6pt 0pt 6pt]{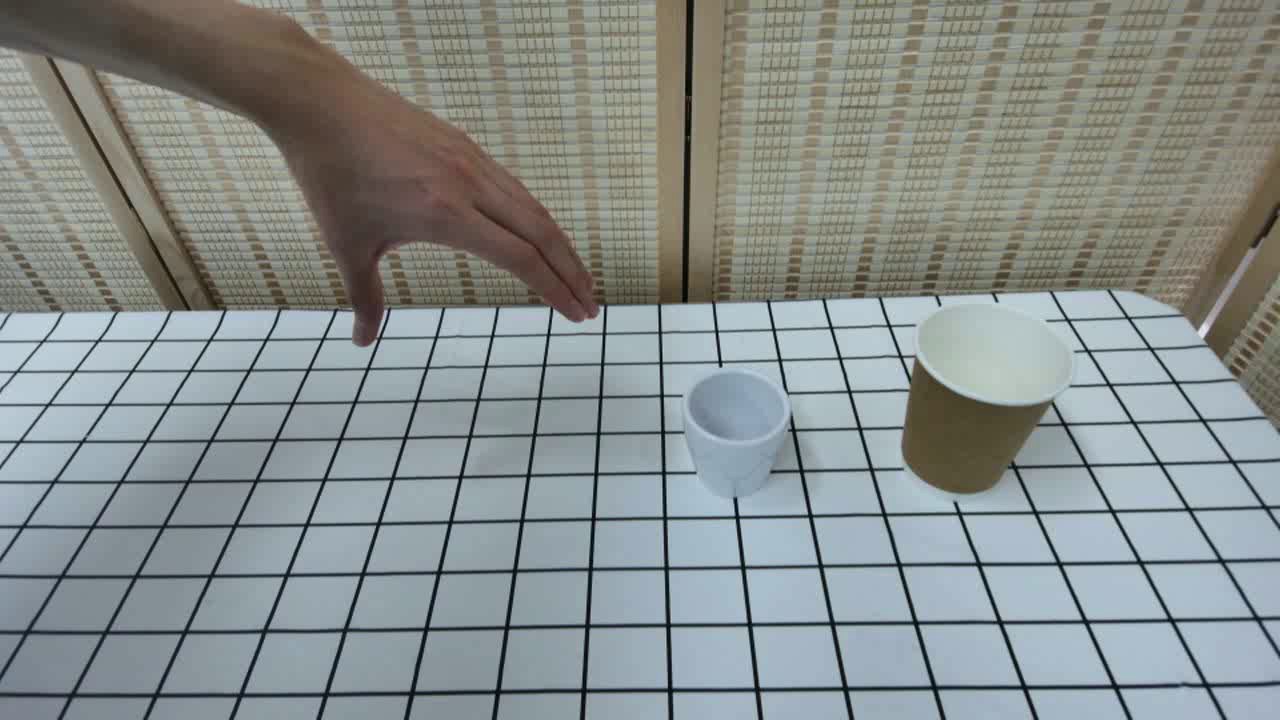} & 
    The hand grasps the plastic cup using the thumb and the index finger. It slowly lifts the cup from the table. It tilts the cup to pour water into the paper cup. It releases the plastic cup. \\

    \hline

    Stack Cups & 
    \includegraphics[width=0.24\textwidth, valign=t, margin=0pt 6pt 0pt 6pt]{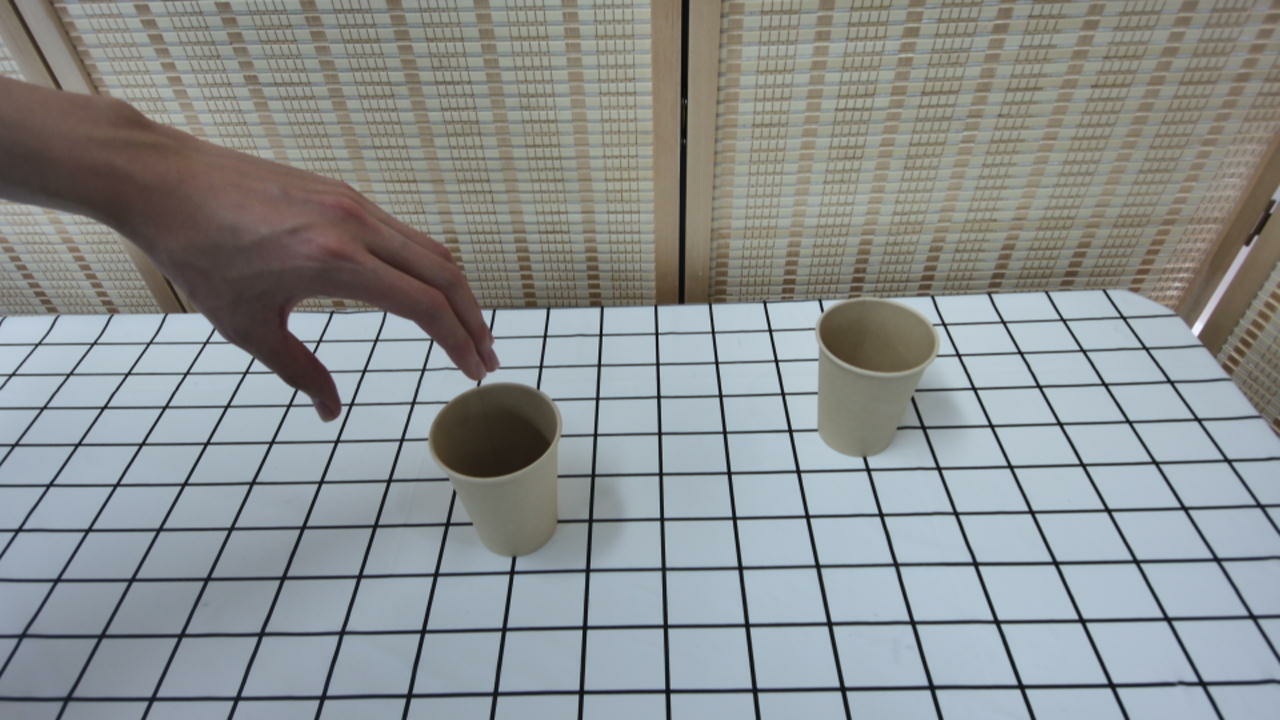} & 
    The hand grasps the left edge of the left cup using the thumb and the index fingers and slowly lifts it, then moves it above the right cup and lowers it to stack inside. \\

    \hline

    Open Drawer & 
    \includegraphics[width=0.24\textwidth, valign=t, margin=0pt 6pt 0pt 6pt]{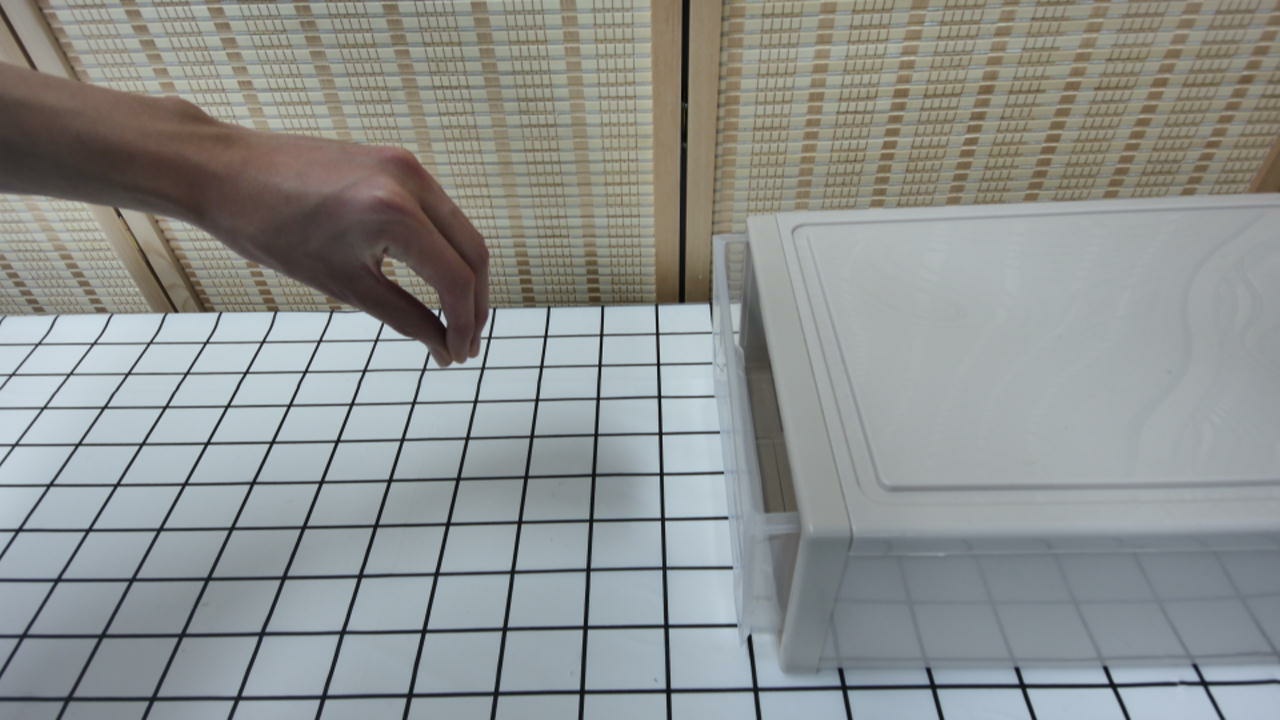} & 
    The hand grasps the left edge of the transparent drawer using the thumb and the index fingers then pulls it leftward causing the drawer to slide open. \\

    \hline

    Put in Drawer & 
    \includegraphics[width=0.24\textwidth, valign=t, margin=0pt 6pt 0pt 6pt]{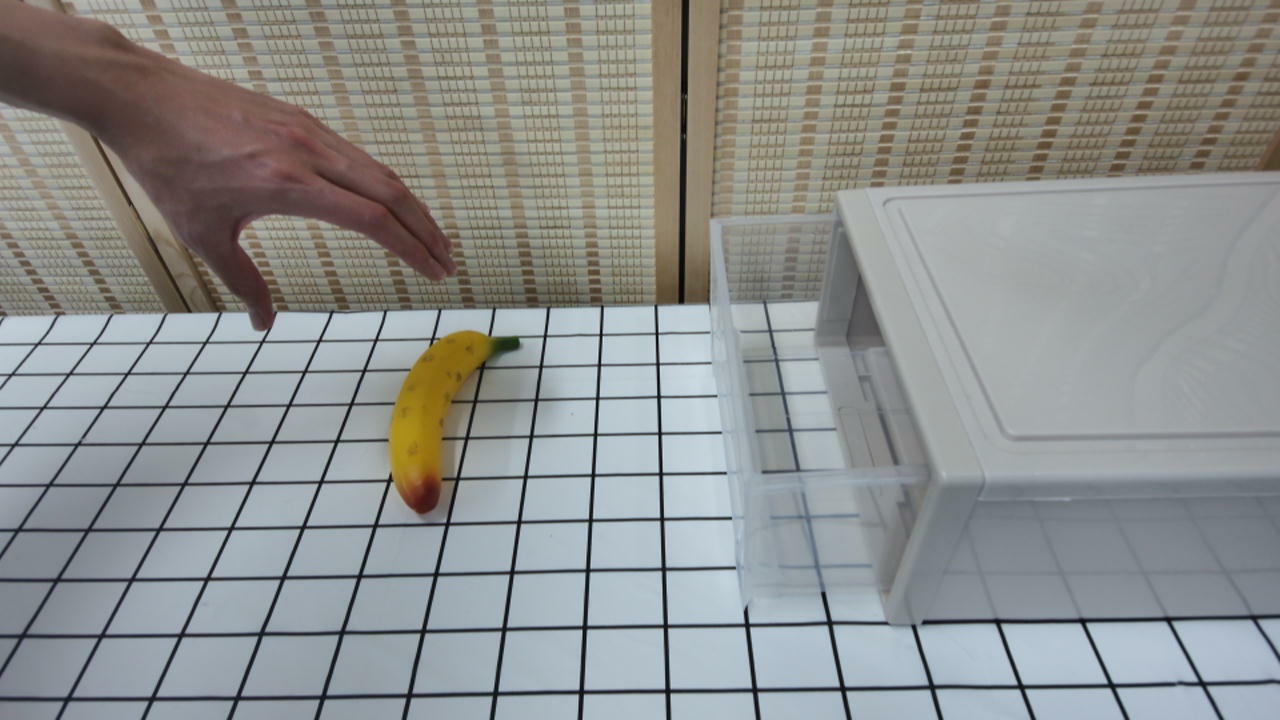} & 
    The hand slowly moves to grasp the banana using the thumb and the index finger, lifts it and places it inside the clear drawer. The hand pushes the drawer into the container until it is fully closed. \\

    \hline

    Fold Towel & 
    \includegraphics[width=0.24\textwidth, valign=t, margin=0pt 6pt 0pt 6pt]{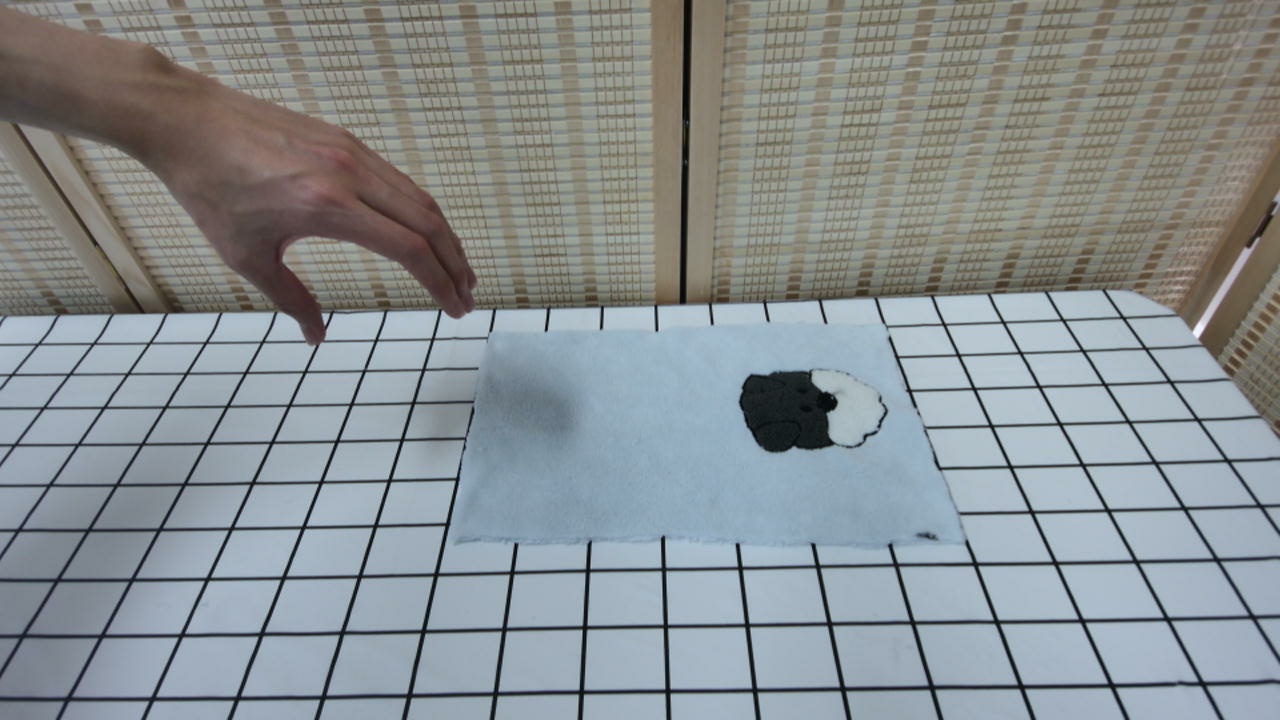} & 
    The hand grasps the left edge using the thumb and the index fingers and lifts it. The hand folds the towel over to the right edge. The hand aligns the edges and presses down to secure the fold. \\
    
    \hline
    \end{tabularx}
\end{table*}

\begin{table*}[htbp]
    
    \centering
    \small 
    
    \setlength{\aboverulesep}{0pt}
    \setlength{\belowrulesep}{0pt}
    \setlength{\tabcolsep}{8pt}    
    \renewcommand{\arraystretch}{1.4} 

    \caption{Overview of human-hand manipulation tasks and their corresponding instruction prompts (continued).}
    \label{tab:human_bi_prompts}	
    \vspace{0.8em}

    \newcolumntype{K}{>{\centering\arraybackslash}p{2.2cm}}
    
    \newcolumntype{M}{>{\centering\arraybackslash}p{0.28\textwidth}}
    
    \renewcommand{\tabularxcolumn}[1]{p{#1}} 
    \newcolumntype{L}{>{\raggedright\arraybackslash\scriptsize}X} 

    \begin{tabularx}{\textwidth}{K | M | L}
    \hline
    \textbf{Task Name} & \textbf{Input Image} & \multicolumn{1}{>{\centering\arraybackslash}X}{\textbf{Task Prompt}} \\
    \hline

    Cook & 
    \includegraphics[width=0.24\textwidth, valign=t, margin=0pt 6pt 0pt 6pt]{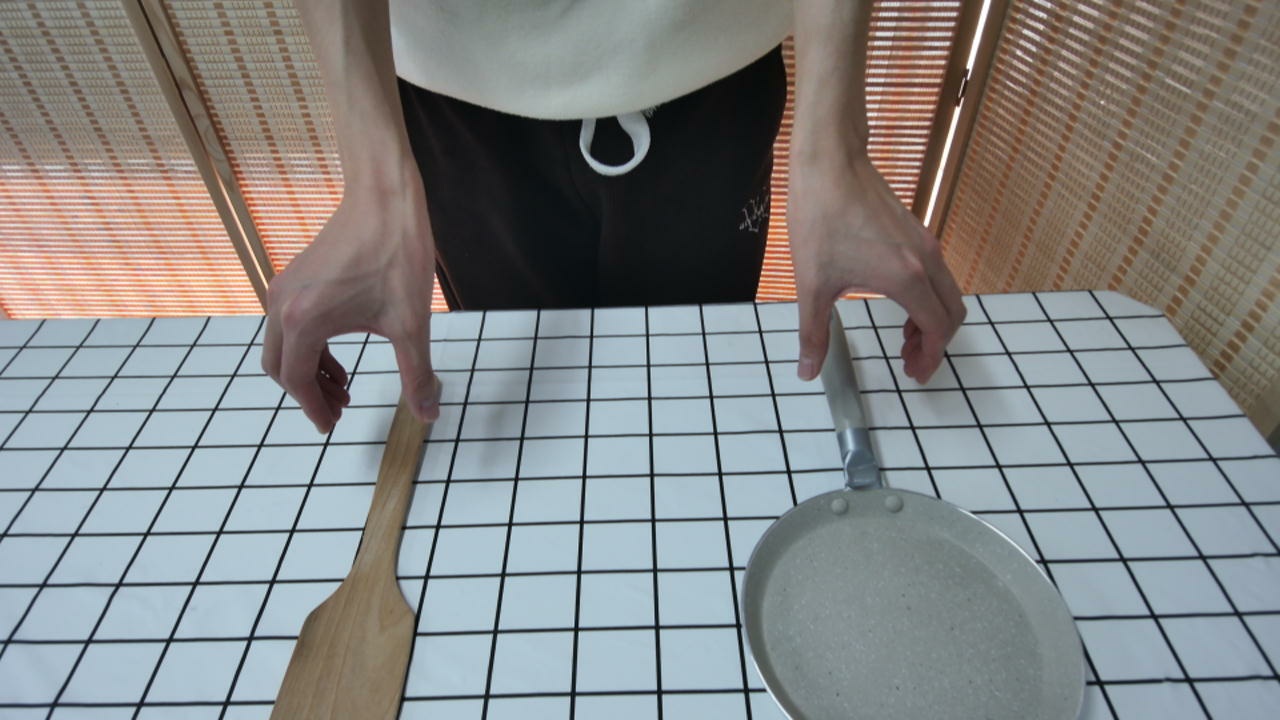} & 
    The person uses their left hand to pick up the wooden spatula and places it into the frying pan, while their right hand holds the handle of the pan and lift it up. They move the spatula around inside the pan briefly as if stirring or scraping, then lift the spatula out and place the wooden spatula and the frying pan back on the table in their original position. \\
    
    \hline

    Lift Large Box & 
    \includegraphics[width=0.24\textwidth, valign=t, margin=0pt 6pt 0pt 6pt]{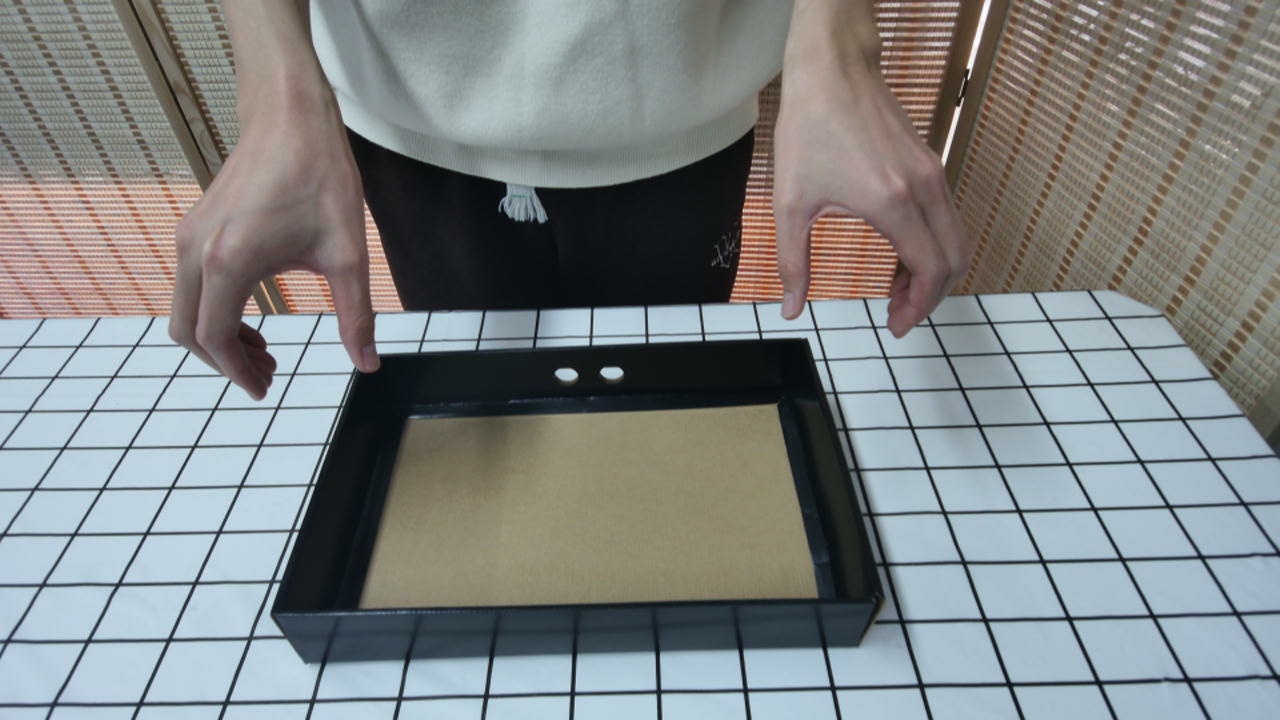} & 
    The person grasps both the left and right edges of the box using the thumb and the index fingers, lifting it vertically off the table. \\
    
    \hline
    \end{tabularx}
\end{table*}

\newpage
\subsection{Robotic Tasks}

To generate instruction prompts for video world models on robotic manipulation tasks, we also leverage the Qwen3-VL-Flash \cite{Qwen3-VL} model to generate concise task-specific descriptions conditioned on the initial observation image and high-level task instructions (e.g., \emph{close drawer}). The resulting descriptions generated by Qwen are summarized in Table~\ref{tab:robot_prompts} and Table~\ref{tab:sim_prompts}.

To further encourage physical consistency of the synthesized videos, we append a standardized set of constraints to the generated instruction: \textit{"Throughout the entire video, the gripper undergoes no structural deformation, only the opening angle of its jaws changes; the rotational movements of the robotic arm joints strictly adhere to its inherent mechanical structure; and the camera perspective remains completely unchanged."} Empirically, we observe that including these constraints improves the quality of videos generated by the world models.

\begin{table*}[htbp]
    
    \centering
    \small 
    
    \setlength{\aboverulesep}{0pt}
    \setlength{\belowrulesep}{0pt}
    \setlength{\tabcolsep}{8pt}    
    \renewcommand{\arraystretch}{1.4} 

    \caption{Overview of real-to-sim robotic manipulation tasks and their corresponding instruction prompts.}
    \label{tab:robot_prompts}	
    \vspace{0.8em}

    \newcolumntype{K}{>{\centering\arraybackslash}p{2.2cm}}
    
    \newcolumntype{M}{>{\centering\arraybackslash}p{0.28\textwidth}}
    
    \renewcommand{\tabularxcolumn}[1]{p{#1}} 
    \newcolumntype{L}{>{\raggedright\arraybackslash\scriptsize}X} 

    \begin{tabularx}{\textwidth}{K | M | L}
    \hline
    \textbf{Task Name} & \textbf{Input Image} & \multicolumn{1}{>{\centering\arraybackslash}X}{\textbf{Task Prompt}} \\
    \hline

    Close Drawer & 
    \includegraphics[width=0.24\textwidth, valign=t, margin=0pt 6pt 0pt 6pt]{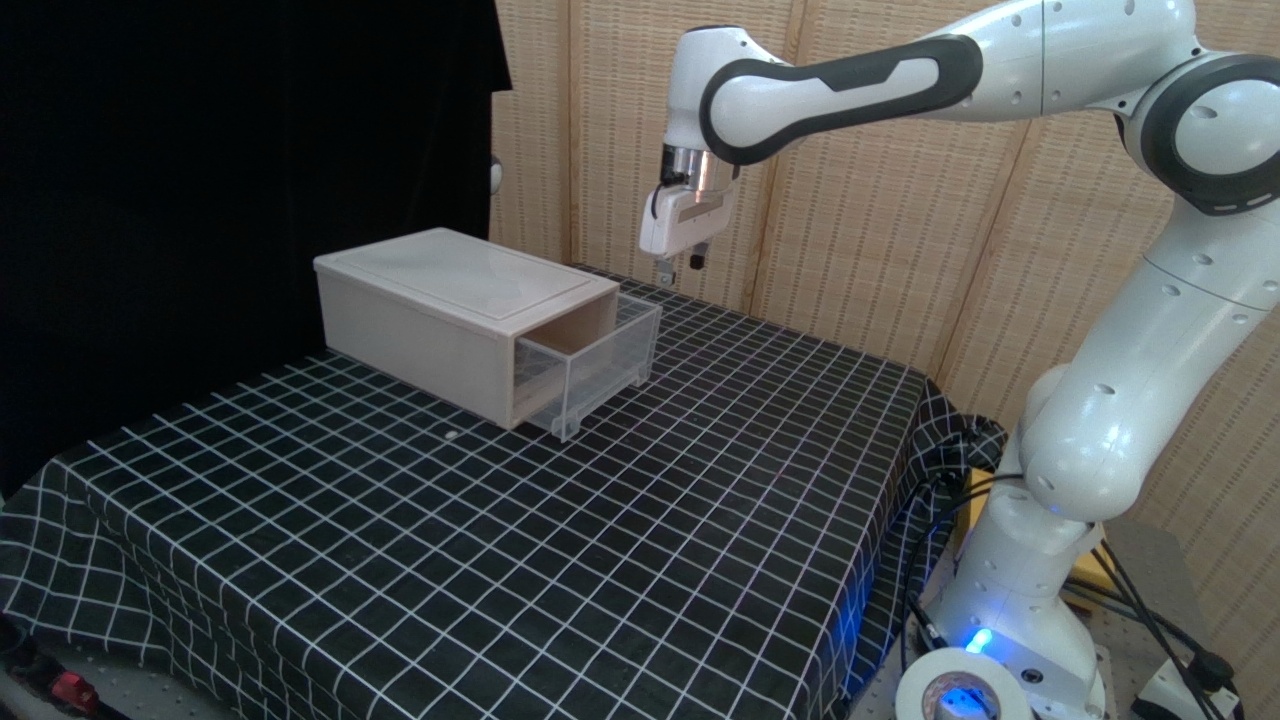} & 
    The robotic arm closes its gripper and pushes the drawer back to its closed position. \\
    
    \hline

    Pick Object & 
    \includegraphics[width=0.24\textwidth, valign=t, margin=0pt 6pt 0pt 6pt]{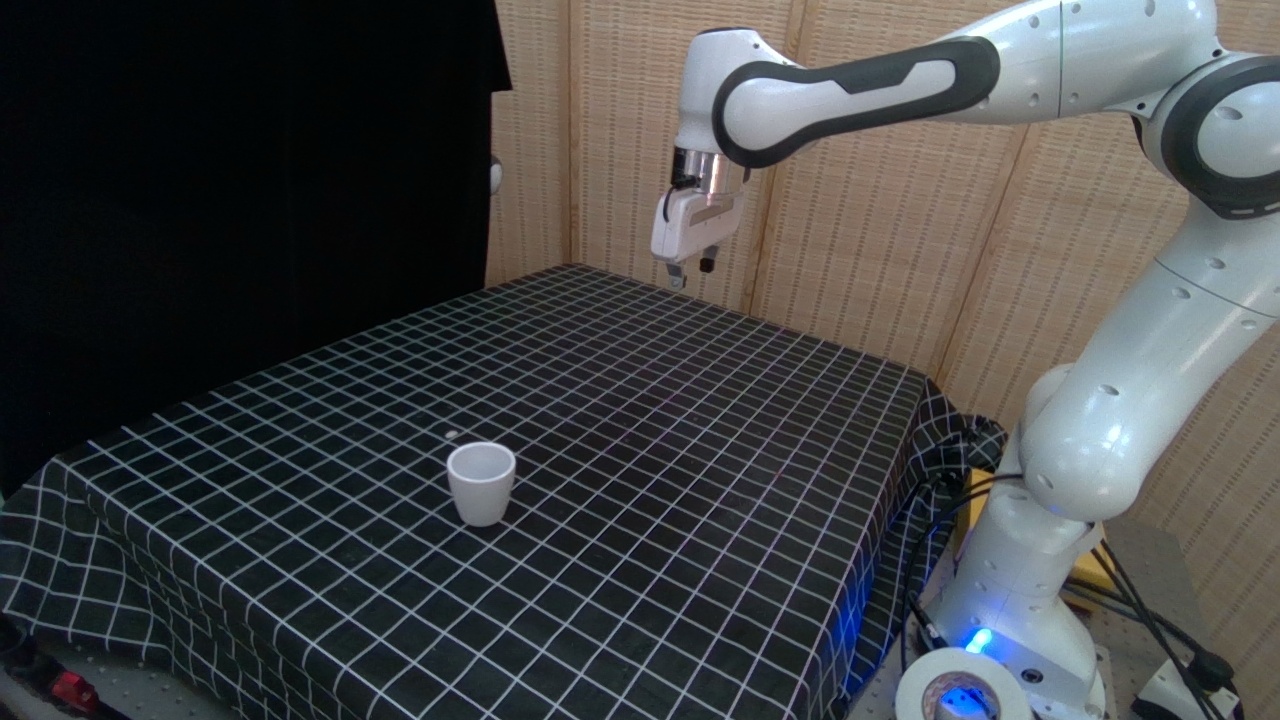} & 
    The robotic arm picks up the white cup from the table surface. \\
    
    \hline

    Push Object & 
    \includegraphics[width=0.24\textwidth, valign=t, margin=0pt 6pt 0pt 6pt]{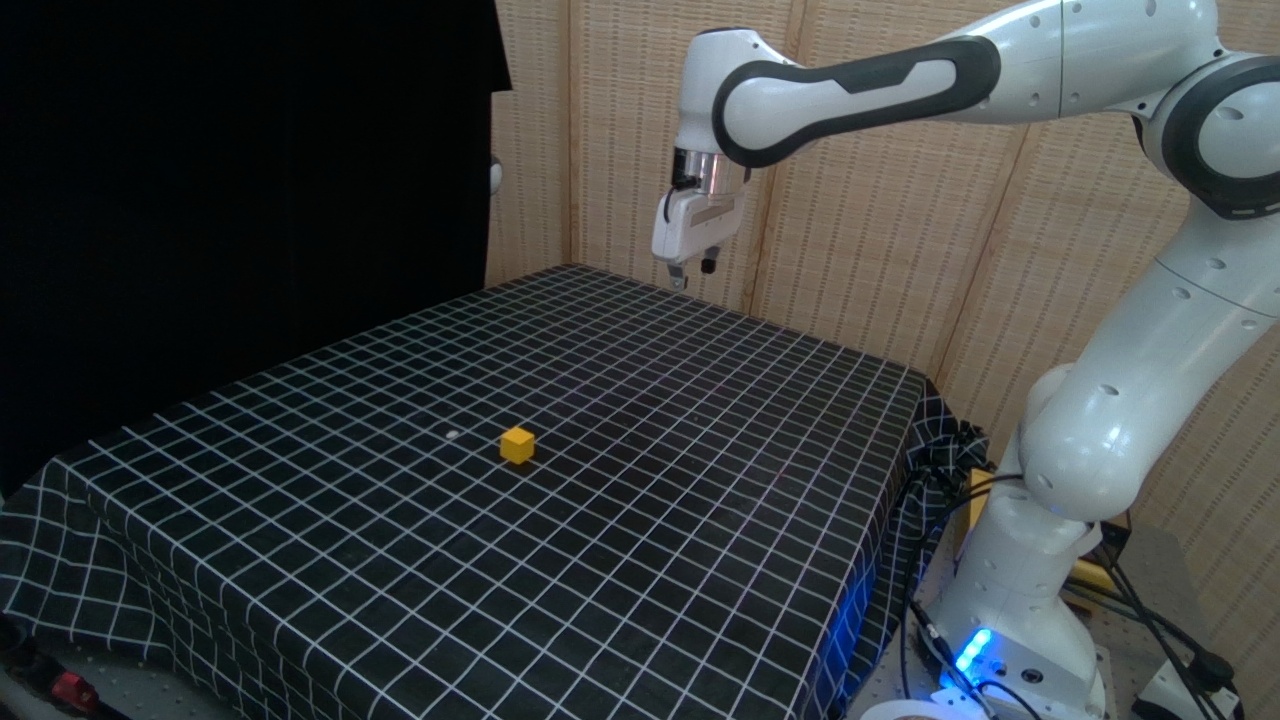} & 
    The robotic arm closes its gripper and pushes the green cube away from the base of the arm for a short distance. \\

    \hline

    Push Button & 
    \includegraphics[width=0.24\textwidth, valign=t, margin=0pt 6pt 0pt 6pt]{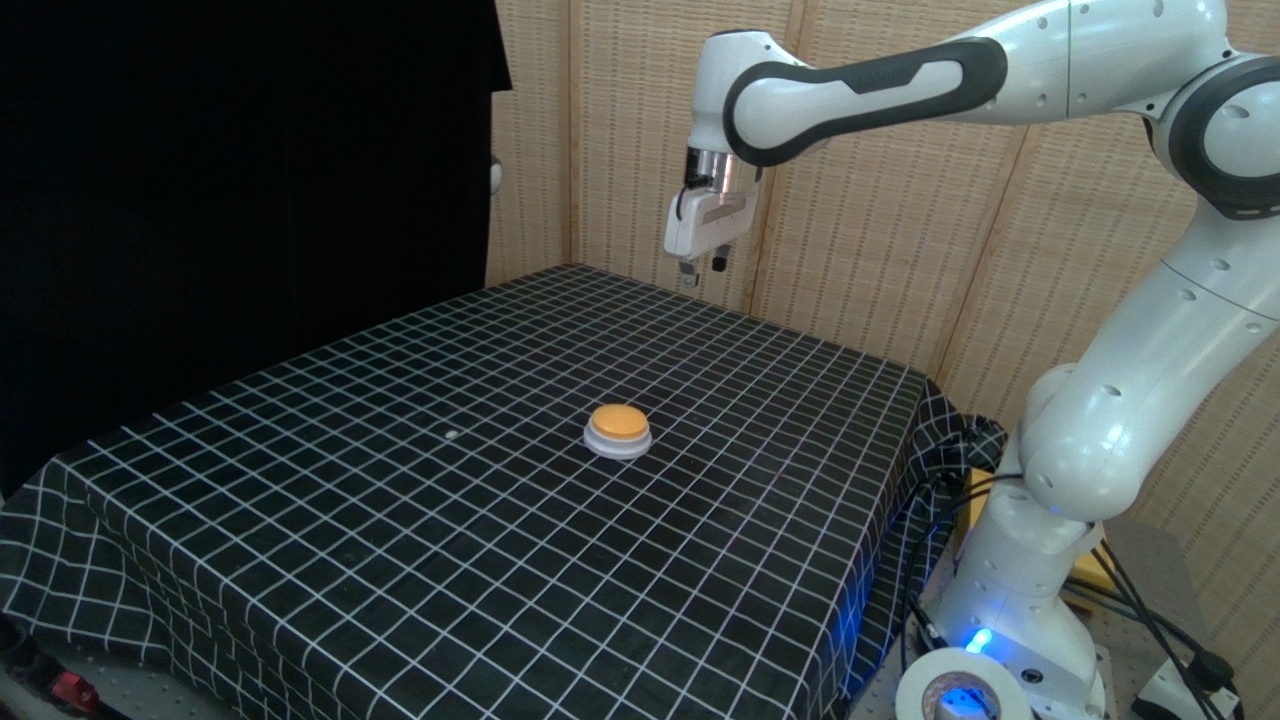} & 
    The robotic arm closes its gripper, then uses the gripper to press the yellow button on the tabletop. \\

    \hline

    Put on Plate & 
    \includegraphics[width=0.24\textwidth, valign=t, margin=0pt 6pt 0pt 6pt]{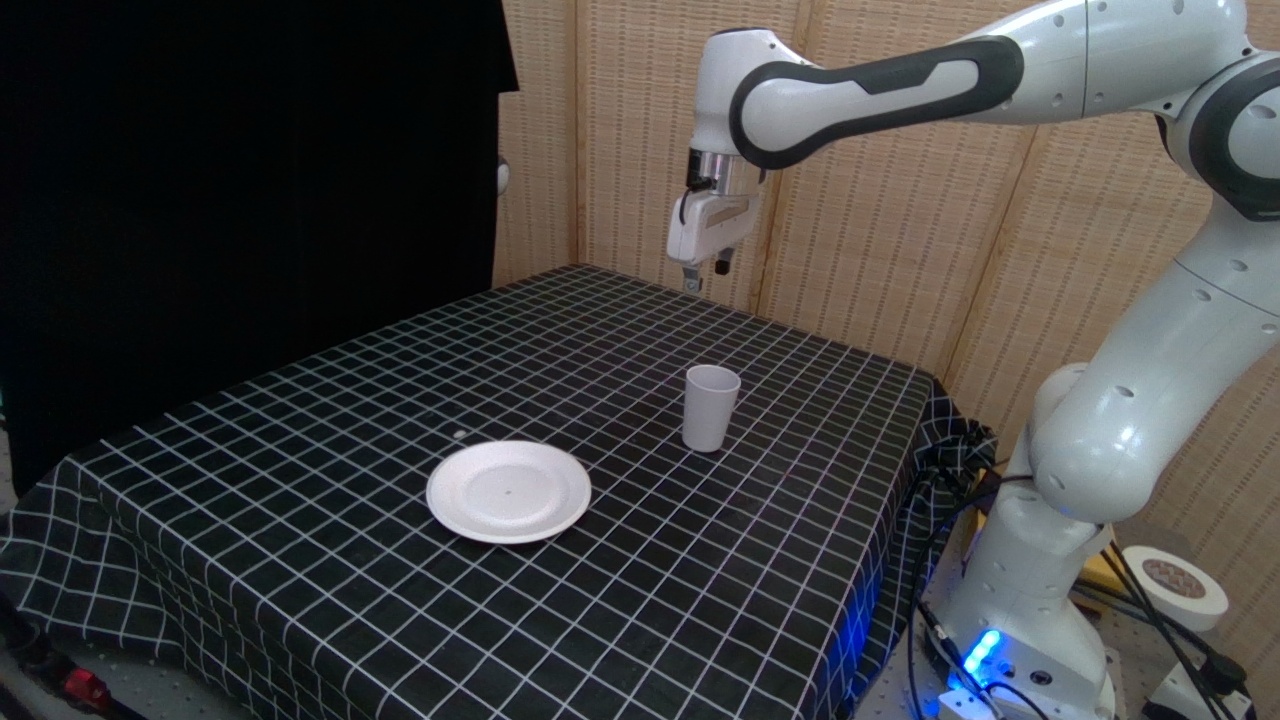} & 
    The robotic arm picks up the white cup from the table and moves its gripper above the plate, then releases the cup placing it on the plate. \\

    \hline

    Discard Trash & 
    \includegraphics[width=0.24\textwidth, valign=t, margin=0pt 6pt 0pt 6pt]{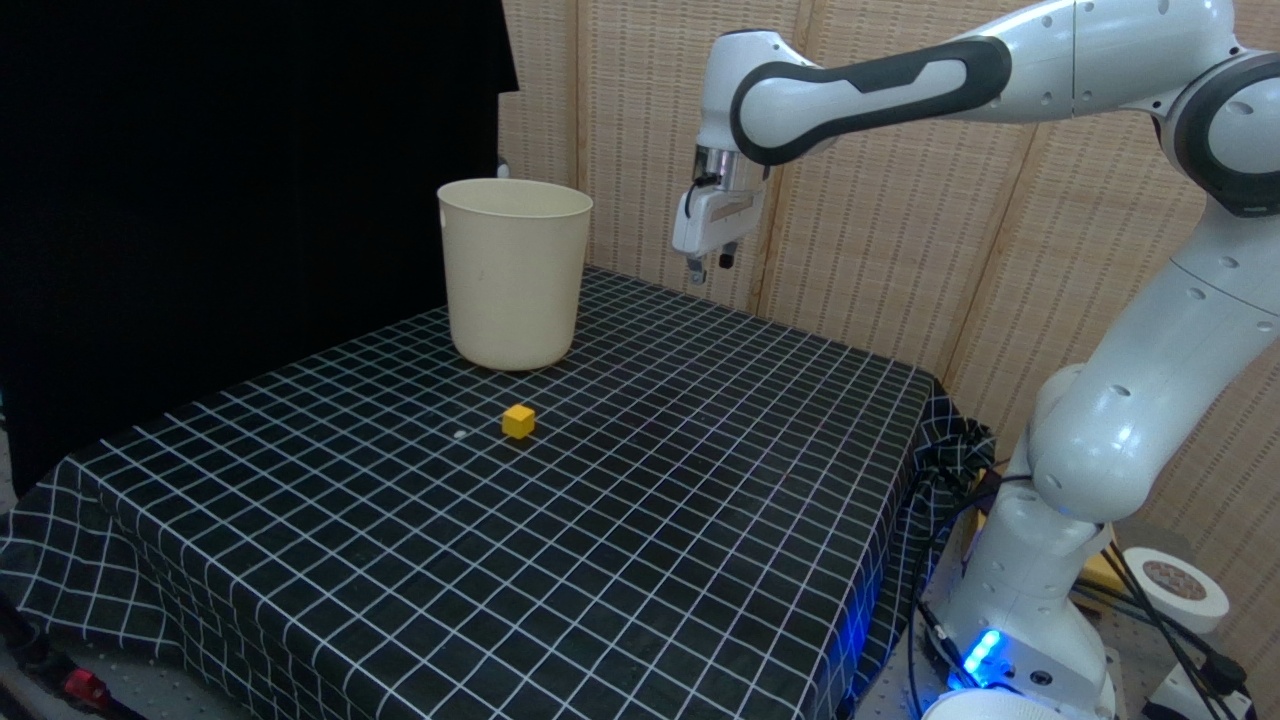} & 
    The robotic arm picks up the green cube from the table and moves its gripper above the trash bin, then releases the gripper to drop the green cube into the trash bin. \\

    \hline

    Pull Object & 
    \includegraphics[width=0.24\textwidth, valign=t, margin=0pt 6pt 0pt 6pt]{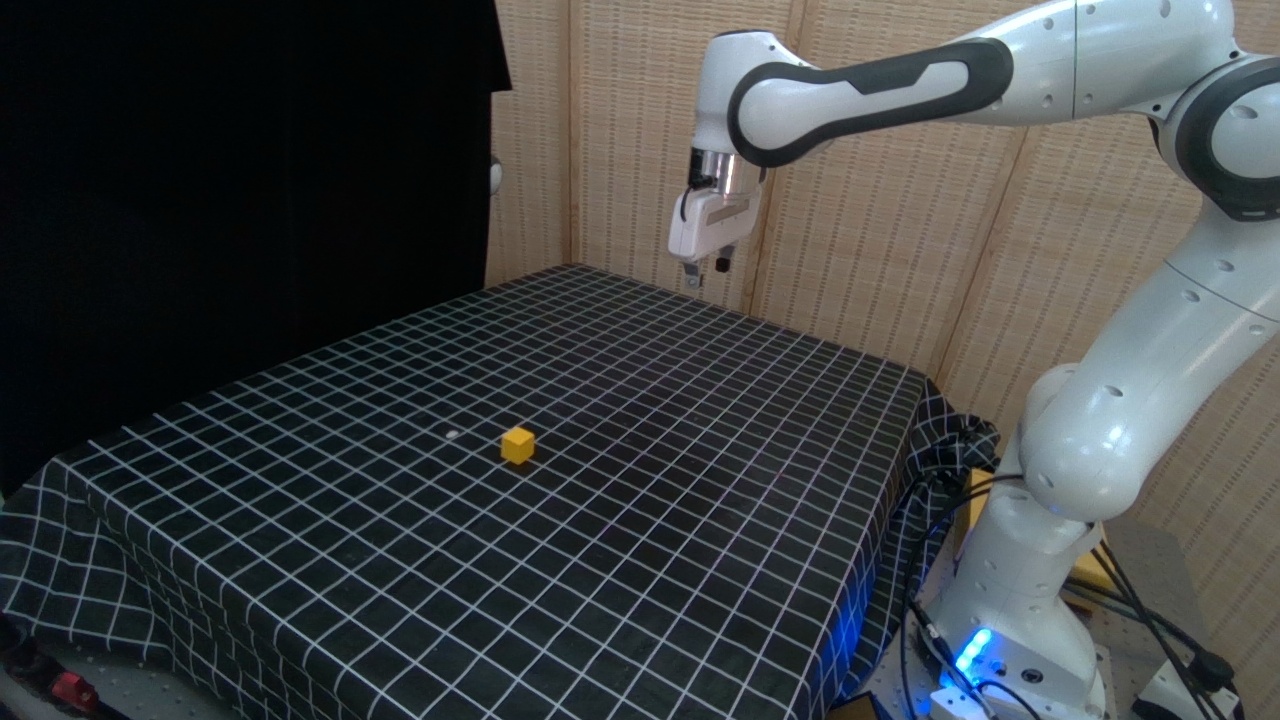} & 
    The robotic arm closes its gripper and pulls the green cube toward the robotic arm base for a short distance. \\

    \hline

    Put in Drawer & 
    \includegraphics[width=0.24\textwidth, valign=t, margin=0pt 6pt 0pt 6pt]{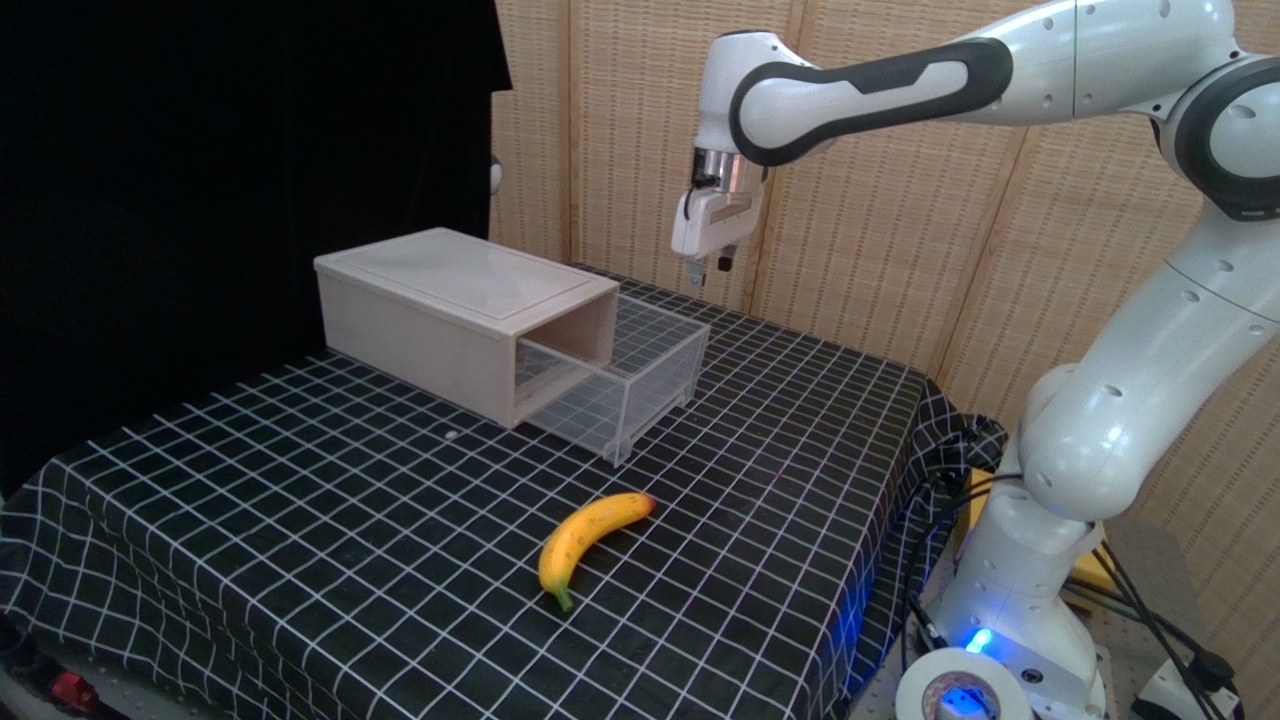} & 
    The robotic arm picks up the yellow banana, moves it above the open drawer, releases the gripper so the banana falls inside, then closes the gripper to push the drawer back to its closed position. \\
    
    \hline
    \end{tabularx}
\end{table*}

\begin{table*}[htbp]
    
    \centering
    \small 
    
    \setlength{\aboverulesep}{0pt}
    \setlength{\belowrulesep}{0pt}
    \setlength{\tabcolsep}{8pt}    
    \renewcommand{\arraystretch}{1.4} 

    \caption{Overview of purely simulated robotic manipulation tasks and their corresponding instruction prompts.}
    \label{tab:sim_prompts}	
    \vspace{0.8em}

    \newcolumntype{K}{>{\centering\arraybackslash}p{2.2cm}}
    
    \newcolumntype{M}{>{\centering\arraybackslash}p{0.28\textwidth}}
    
    \renewcommand{\tabularxcolumn}[1]{p{#1}} 
    \newcolumntype{L}{>{\raggedright\arraybackslash\scriptsize}X} 

    \begin{tabularx}{\textwidth}{K | M | L}
    \hline
    \textbf{Task Name} & \textbf{Input Image} & \multicolumn{1}{>{\centering\arraybackslash}X}{\textbf{Task Prompt}} \\
    \hline

    Close Drawer & 
    \includegraphics[width=0.24\textwidth, valign=t, margin=0pt 6pt 0pt 6pt]{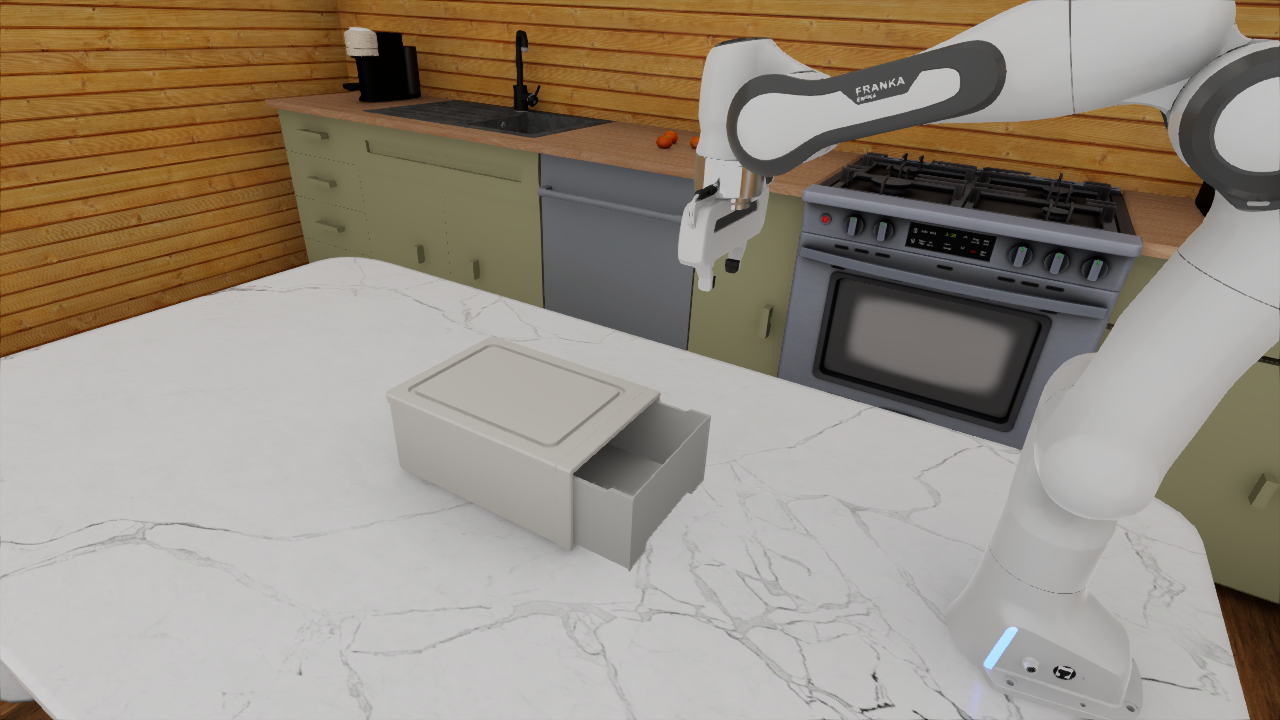} & 
    The robotic arm closes its gripper and pushes the drawer back to its closed position. \\
    
    \hline
    
    Push Button & 
    \includegraphics[width=0.24\textwidth, valign=t, margin=0pt 6pt 0pt 6pt]{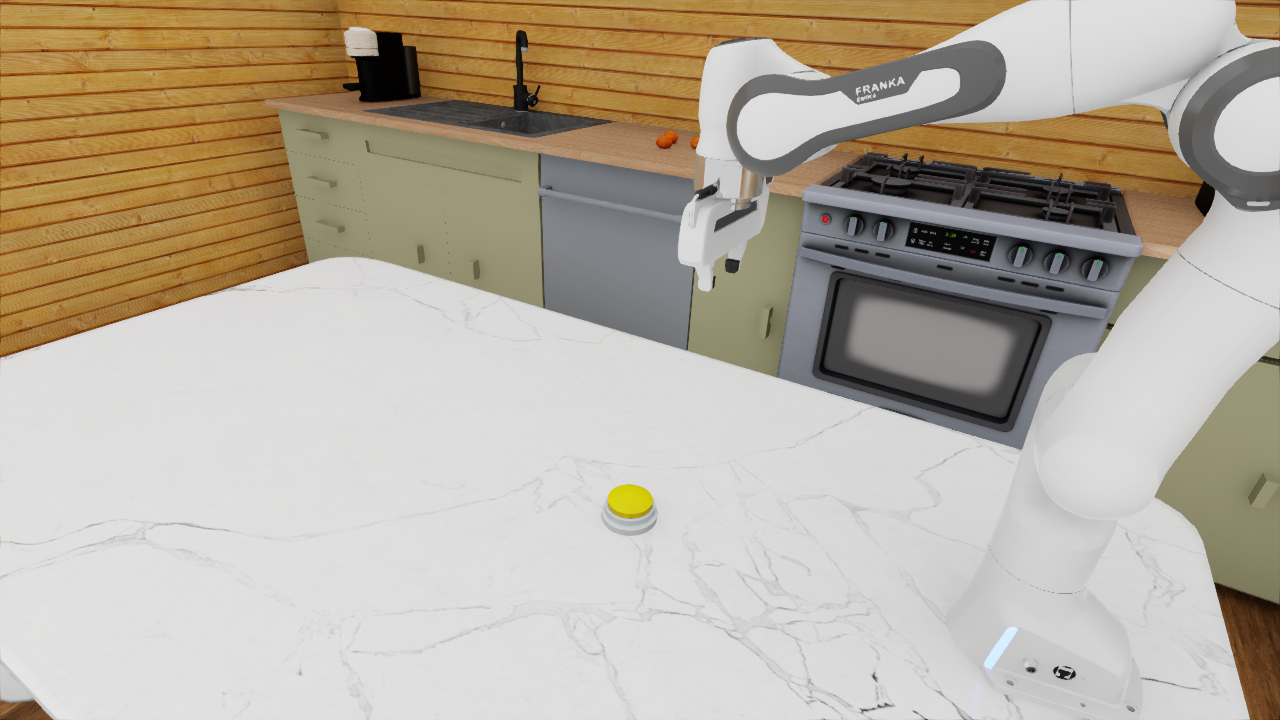} & 
    The robotic arm closes its gripper, then uses the gripper to press the yellow button on the tabletop. \\

    \hline
    Cut Sausage & 
    \includegraphics[width=0.24\textwidth, valign=t, margin=0pt 6pt 0pt 6pt]{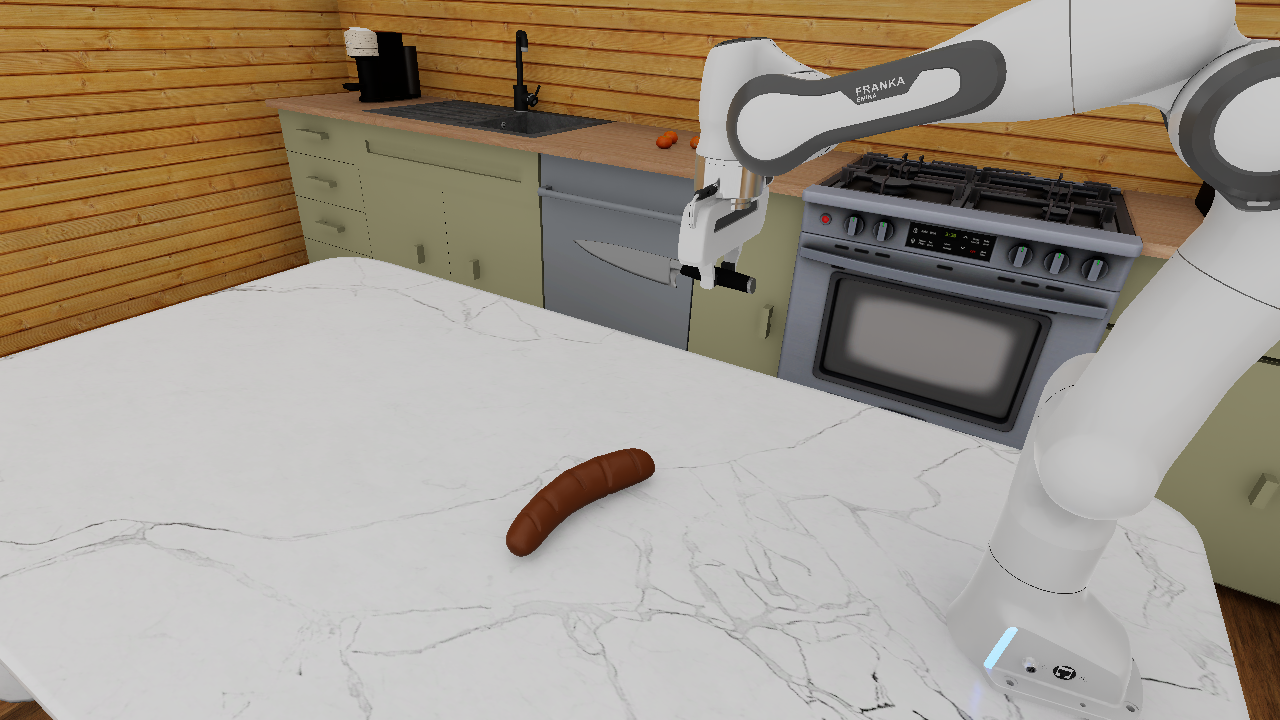} & 
    The robotic arm uses a knife to cut the sausage on the table. \\
    
    \hline
    
    Turn Off Faucet & 
    \includegraphics[width=0.24\textwidth, valign=t, margin=0pt 6pt 0pt 6pt]{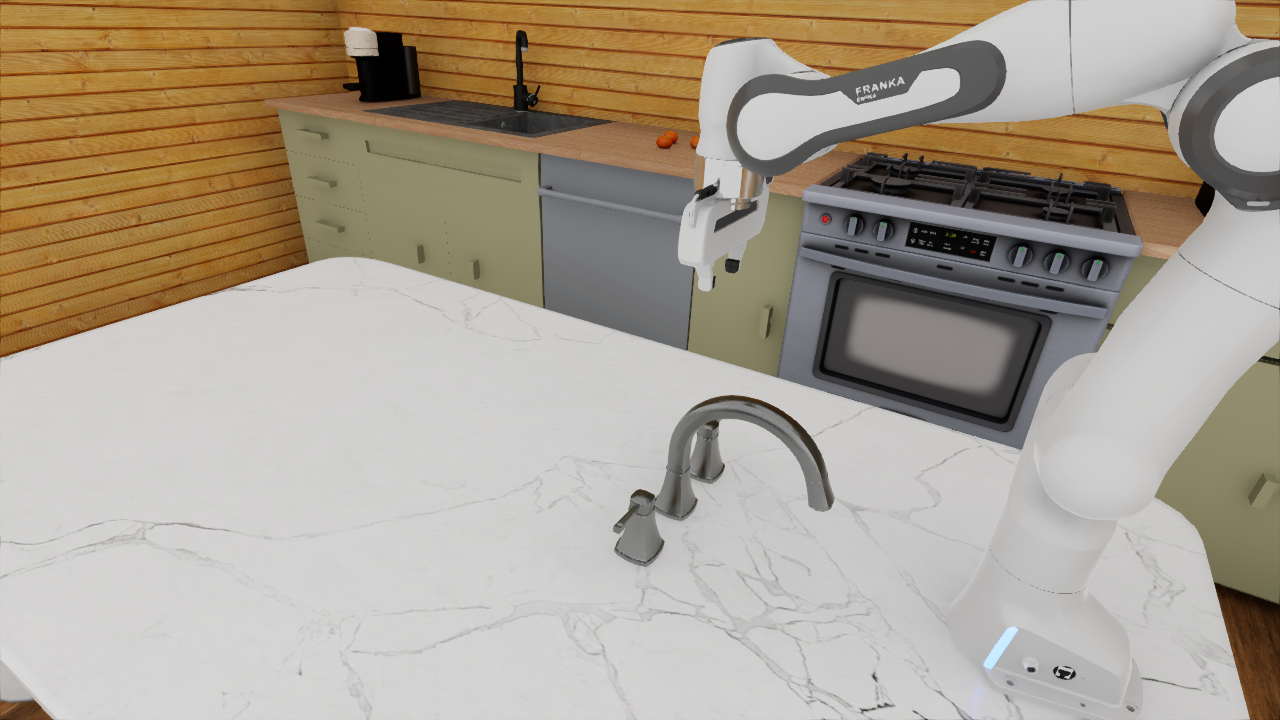} & 
    The robotic arm grasps the lever handle on the left side of the faucet, then rotates it inward to turn off the water. \\
    
    \hline

    Assemble Burger & 
    \includegraphics[width=0.24\textwidth, valign=t, margin=0pt 6pt 0pt 6pt]{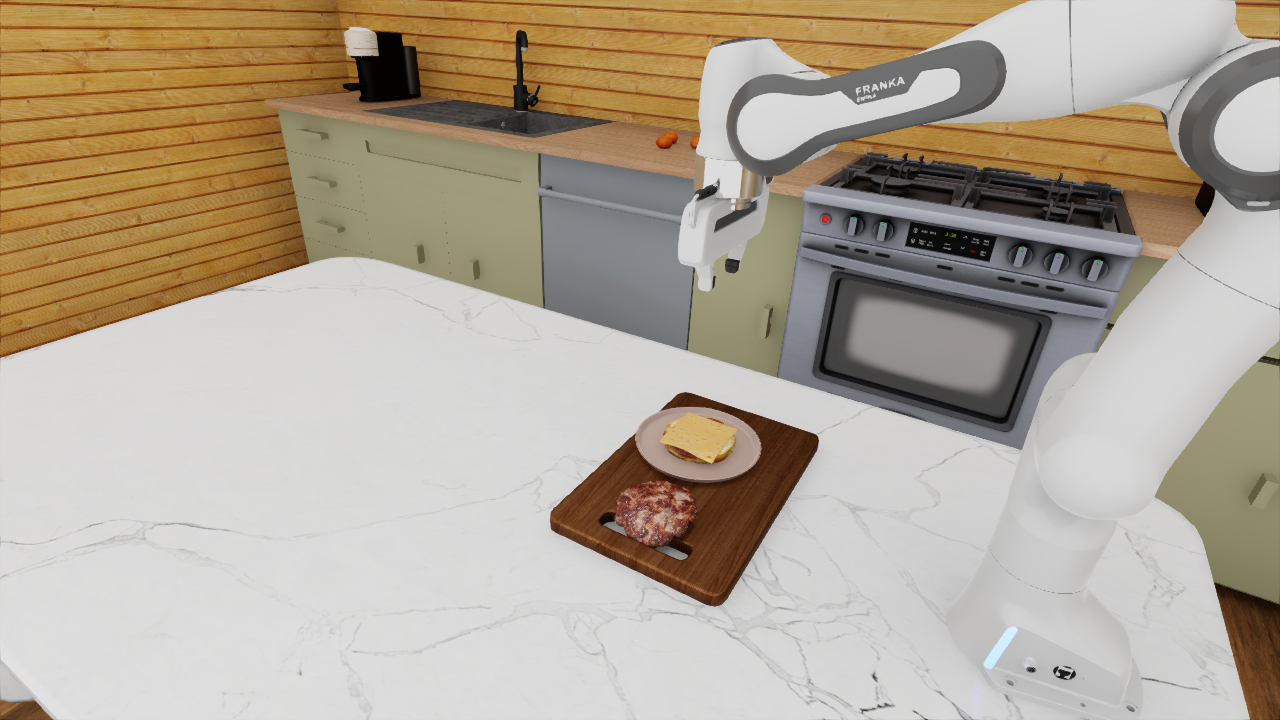} & 
    The robotic arm first pushes the meat patty from the left side of the cutting board to its edge, then picks it up and places it on top of the cheese in the plate located on the right side of the cutting board. \\

    \hline

    Fold Clothes & 
    \includegraphics[width=0.24\textwidth, valign=t, margin=0pt 6pt 0pt 6pt]{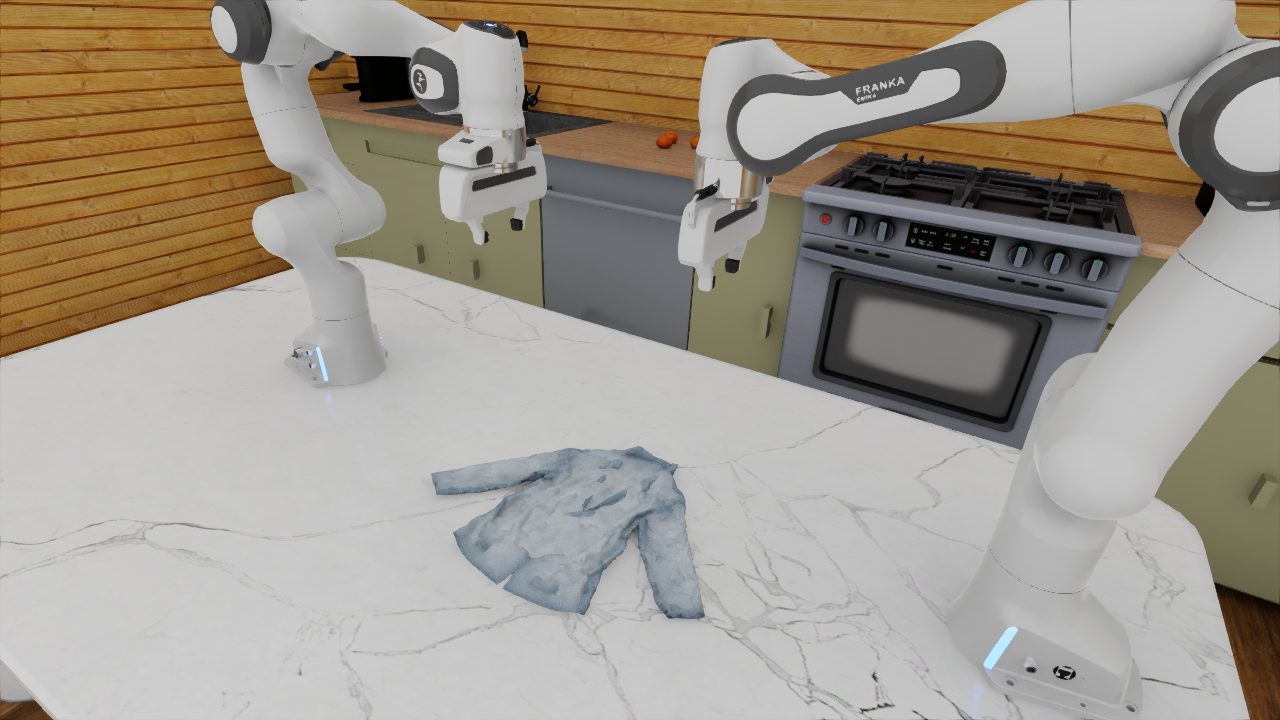} & 
    The left arm grasps the cuff of the left sleeve, while the right arm grasps the cuff of the right sleeve. Both arms then lift and fold the sleeves inward toward the center of the shirt: the left arm places the left cuff onto the left side of the shirt's body, and the right arm places the right cuff onto the right side of the body, aligning them neatly along the torso. \\

    \hline

    \end{tabularx}
\end{table*}

\newpage
\section{Implementation Details of PAI-Bench Domain Score}
\label{sec:pai_domain_implementation}

To facilitate evaluation within the PAI-Bench framework, we design targeted VQA suites for human-hand tasks and robotic tasks, respectively. Within each suite, only the first question is task-dependent and varies across scenarios, while the remaining four questions are identical across all tasks.

\vspace{1.5em}

\subsection{VQA Pairs for Human-Hand Tasks}

\noindent \textbf{Question 1:} Does the human successfully complete the task: \{The task description, such as the human hand grasping and lifting the banana from the table surface.\} \\
\textbf{options:} (A) \textbf{yes} \quad (B) no \quad (C) unclear \\
\textbf{ground truth:} \textbf{A}

\vspace{1.2em}

\noindent \textbf{Question 2:} Does the human hand make physical contact with the target object? \\
\textbf{options:} (A) \textbf{yes} \quad (B) no \quad (C) unclear \\
\textbf{ground truth:} \textbf{A}

\vspace{1.2em}

\noindent \textbf{Question 3:} Does the human hand maintain anatomically plausible hand structure throughout the video (no impossible bends/broken fingers/extra joints)? \\
\textbf{options:} (A) \textbf{yes} \quad (B) no \quad (C) unclear \\
\textbf{ground truth:} \textbf{A}

\vspace{1.2em}

\noindent \textbf{Question 4:} Do all task-relevant objects maintain their structural integrity without undergoing physically implausible deformations? \\
\textbf{options:} (A) \textbf{yes} \quad (B) no \quad (C) unclear \\
\textbf{ground truth:} \textbf{A}

\vspace{1.2em}

\noindent \textbf{Question 5:} Do all task-relevant objects exhibit physically plausible motions, without any sudden spatial displacement? \\
\textbf{options:} (A) \textbf{yes} \quad (B) no \quad (C) unclear \\
\textbf{ground truth:} \textbf{A}

\subsection{VQA Pairs for Robotic Tasks}

\noindent \textbf{Question 1:} Does the robot successfully complete the task: \{The task description, such as the robotic arm picks up the small yellow cube from the tabletop.\} \\
\textbf{options:} (A) \textbf{yes} \quad (B) no \quad (C) unclear \\
\textbf{ground truth:} \textbf{A}

\vspace{1.2em}

\noindent \textbf{Question 2:} Does the robot gripper/hand make physical contact with the target object? \\
\textbf{options:} (A) \textbf{yes} \quad (B) no \quad (C) unclear \\
\textbf{ground truth:} \textbf{A}

\vspace{1.2em}

\noindent \textbf{Question 3:} Does the robotic system (arm and gripper) maintain structural integrity and a physically plausible configuration throughout the video, with no deformation of rigid links or gripper, and realistic joint rotations? \\
\textbf{options:} (A) \textbf{yes} \quad (B) no \quad (C) unclear \\
\textbf{ground truth:} \textbf{A}

\vspace{1.2em}

\noindent \textbf{Question 4:} Do all task-relevant objects maintain their structural integrity without undergoing physically implausible deformations? \\
\textbf{options:} (A) \textbf{yes} \quad (B) no \quad (C) unclear \\
\textbf{ground truth:} \textbf{A}

\vspace{1.2em}

\noindent \textbf{Question 5:} Do all task-relevant objects exhibit physically plausible motions, without any sudden spatial displacement? \\
\textbf{options:} (A) \textbf{yes} \quad (B) no \quad (C) unclear \\
\textbf{ground truth:} \textbf{A}

\end{document}